\documentclass{article}

\usepackage{iclr2027_conference,times}
\iclrfinalcopy
\PassOptionsToPackage{numbers, compress}{natbib}

\usepackage[utf8]{inputenc}
\usepackage[T1]{fontenc}
\usepackage[english]{babel}
\usepackage{microtype}
\usepackage{xcolor}
\usepackage{hyperref}
\usepackage{url}
\usepackage{graphicx}
\usepackage{float}
\usepackage{caption}
\usepackage{placeins}
\usepackage{wrapfig}
\usepackage{booktabs}
\usepackage{enumitem}
\usepackage{amsmath}
\usepackage{amsfonts}
\usepackage{amssymb}
\usepackage{amsthm}
\usepackage[capitalize,noabbrev]{cleveref}
\usepackage{algorithm}
\usepackage{algpseudocode}

\theoremstyle{definition}
\newtheorem{definition}{Definition}[section]

\newcommand{\E}{\mathbb{E}}
\newcommand{\Pset}{\mathcal{P}}
\newcommand{\V}[1]{\boldsymbol{\mathbf{#1}}}
\newcommand{\given}{\,|\,}

\DeclareMathOperator{\nie}{NIE}
\DeclareMathOperator{\NIE}{NIE}
\DeclareMathOperator{\pie}{PIE}
\DeclareMathOperator{\WINIE}{WINIE}
\DeclareMathOperator{\WIPIE}{WIPIE}
\DeclareMathOperator{\WISE}{WISE}

\title{Slaying the Hydra: Interaction-Aware Circuit Discovery in Language Models}

\author{%
  \textbf{Sankaran Vaidyanathan} $^{1,2}$\thanks{Co-first authorship. A full contribution statement is available at the end of the paper. Correspondence to \texttt{sankaranv@cs.umass.edu} and \texttt{rfl.urbaniak@gmail.com}.} \quad
  \textbf{Rafal Urbaniak} $^{1}$\footnotemark[1] \quad
  \textbf{Emily Bunnapradist} $^{1}$ \\
  \textbf{Michelangelo Naim} $^{1}$ \quad
  \textbf{Daniel Waxman} $^{1,3}$ \\[6pt]
  \normalfont
  $^{1}$ Basis Research Institute \quad
  $^{2}$ University of Massachusetts Amherst \quad
  $^{3}$ MIT
  }

\begin{document}
\maketitle

\begin{abstract}
Localizing behavior to individual components of a language model is a central goal of mechanistic interpretability. However, scoring components one at a time misses context-dependent effects: a \textit{primary} component can inhibit the activation of a \textit{backup}, leading to issues with ranking components. \emph{Actual causality} studies the structure of such interactions via \textit{witnesses}: variables that provide contextual information to resolve interaction terms. However, estimation with witnesses typically requires combinatorial enumeration and is infeasible in practice. We introduce the \textit{witness-integrated set effect (WISE)}, a family of causal estimands that build on the witness mechanism while taking expectations over \textit{sets} of causes and witnesses to remain computationally feasible. Building on this approach, we introduce \textsc{JuntaLearner}, a gradient-based circuit discovery method that learns to rank components by their causal impact across varying-sized sets of components and witnesses. Alongside faithfulness metrics, we introduce measures of necessity and task specificity, and the \textit{circuit recognition score (CRS)} to summarize each metric across circuit sizes while emphasizing effects achieved by small circuits. Across tasks and models of increasing size, \textsc{JuntaLearner} achieves higher mean CRS compared to attribution baselines on all metrics. Since its cost does not grow with the number of candidate components, \textsc{JuntaLearner} scales to large models while accounting for set-level interactions and avoiding first-order approximations.
\end{abstract}

\section{Introduction}
\label{sec:intro}

Localizing behavior to individual components of a language model is a fundamental goal of mechanistic interpretability \citep{olah2020zoom}. The standard measures of each component's causal contribution are the \textit{natural indirect effect (NIE)} \citep{pearl2001direct} and \textit{pure indirect effect (PIE)} \citep{vanderweele2013ThreewayDecomposition}, often referred to as \textit{noising} and \textit{denoising}, respectively \citep{heimersheim2024patching}. Activation patching \citep{vig2020genderbias,geiger2021causal} computes these quantities exactly, and first-order gradient-based
approximations \citep{hanna2024have,syed2024attribution} have made them applicable at scale \citep{ameisen2025circuit,marks2025sparse}.

However, individual NIE scores are systematically distorted by \textit{interactions} between components; genuine causes can be missed and individual effects need not compose into circuit effects \citep{mueller2024missed}. The canonical example is \textit{preemption}: a primary component masks a backup that takes over once the primary is patched \citep{wang2023interpretability}, a form of self-repair that \citet{mcgrath2023hydra} call the \textit{Hydra effect}, after the monster that grows new heads for each one cut off \citep{hesiod_theogony}.
Interactions can also be \textit{superadditive} or \textit{subadditive}, with joint effects above or below the sum of individual effects \citep{vanderweele2014tutorial}.
These phenomena reflect a combinatorial structure: the joint effect of any component set includes cross-interactions for every combination of its members, which per-component scores cannot separate \citep{vaidyanathan2026curse}.

These failures leave circuit localization caught between Scylla and Charybdis. On one side are per-component scores that ignore interactions. Exact per-component patching needs a forward pass for every component on every prompt, so its cost grows with the model, and it still misses joint effects. First-order approximations are cheaper but discard interactions altogether. On the other side, exhaustive set-level scoring captures interactions but grows exponentially with the number of components, which makes it infeasible for real models. Prior work escapes the search by using prior knowledge to restrict which interactions are considered, either searching for backups to an assumed primary circuit
\citep{wang2023interpretability,gong2026conditionalcoablation} or predefining the interaction structure \citep{chen2025rethinking}. What is missing is a method that accounts for interactions at feasible cost.

\begin{figure}[t]
\centering
\includegraphics[width=\linewidth]{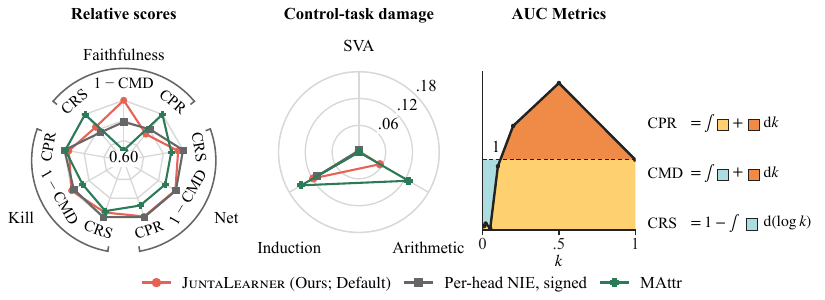}
\caption{\textbf{Interactions and evaluation on GPT-2/IOI.}
\textbf{Left:} performance of \textsc{JuntaLearner}, compared to the signed per-head NIE and MAttr, across a combination of our metrics.
\textbf{Center:} as some task (IOI) is successfully ``killed'' (by removing the estimated $10$-head IOI circuit), other tasks may also take collateral damage; this is captured in our ``net'' metric.
\textbf{Right:} MAttr faithfulness on GPT-2/IOI incurs significant ``overshoot''; CPR rewards this overshoot, while CMD penalizes it. Meanwhile, CRS rewards early discovery by measuring area under the log-curve.}
\label{fig:overview}
\end{figure}

In this paper, we make the following contributions:

\begin{enumerate}[label=(\alph*), leftmargin=0.65cm]

\item \textit{Witness-Integrated Set Effects (WISE)}, a family of set-level causal estimands parameterized by an effect measure such as NIE, PIE or their soft minimum. WISE scores components by averaging causal effects over \textit{sets} in which they interact, while blocking self-repair from outside the set via \textit{witness pinning}, which holds specific downstream sites at their clean values. Witnesses originate in actual causality \citep{halpern2016actual}; we generalize them to the causal mediation setting.

\item \textsc{JuntaLearner}, a scalable circuit discovery method based on WISE maximization over a learned distribution of candidate sets, through a differentiable Gumbel-top-$k$ relaxation with a dynamically annealed set size. A single run produces a component ranking from which circuits of different sizes can be extracted.

\item A set of circuit evaluation metrics that expand on the notion of circuit quality beyond faithfulness: \textit{kill}, \textit{net}, and the \emph{circuit recognition score (CRS)}. Kill measures necessity as the task-performance loss when a circuit is patched, while net measures task specificity by penalizing the collateral damage that patching one task's circuit does to other tasks. CRS summarizes each metric across circuit sizes, rewarding early recognition \citep{truchon2007bedroc} by giving greater weight to high effects achieved by smaller circuits.

\item On five tasks across GPT-2 small, Qwen3-1.7B and Llama-3.1-8B, \textsc{JuntaLearner} outperforms attribution baselines including EAP-IG \citep{hanna2024have} and MAttr \citep{arora2026matryoshka} on kill and net, and performs at least as well as exact per-head NIE. On the public test split of the MIB Circuit
Localization track \citep{mueller2025mib}, a single untuned run of \textsc{JuntaLearner} beats MAttr's released circuits on kill, net, and the CMD metric for faithfulness that MIB uses to penalize overshooting the original model's performance.

\end{enumerate}

\section{Witness-Integrated Causal Effects}
\label{sec:estimand}

\begin{figure}[t]
\centering
\includegraphics[width=\linewidth]{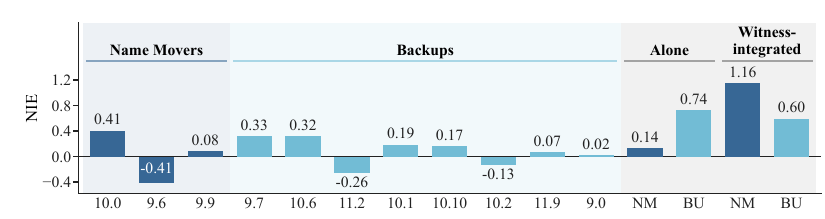}
\caption{\textbf{Witness interactions in the GPT-2 IOI circuit.} The Name Mover group plays a primary role in the GPT-2 IOI circuit, but its backups obscure this role under the standard NIE; witness-integration recovers the intuitive result that the primaries (NM) rank above the backups (BU).}
\label{fig:estimand-sets}
\end{figure}

Counterfactual estimands can give unintuitive answers when causes are redundant, a problem studied at length in actual causality \citep{halpern2016actual}. The standard example is \textit{preemption} \citep{Hall2001TwoCO}: Suzy and Billy both throw rocks at a bottle, and Suzy's arrives first and shatters it. Had she not thrown, Billy's rock would have shattered the bottle anyway, so her throw has no counterfactual effect even though it broke the bottle. Language models show the same pattern \citep{mueller2024missed}: GPT-2's IOI circuit has backup name mover heads that were only active once the primary name movers were ablated \citep{wang2023interpretability}, and ablating a layer can trigger \textit{self-repair} in the subsequent layers \citep{mcgrath2023hydra}.

\paragraph{Indirect Effects}
Take a factual prompt $a$, an alternative prompt $a'$, and a set $\V{M}$ of model components, the \textit{mediators} \citep{steen2017flexible}. The natural indirect effect (NIE) \citep{pearl2001direct} asks how much of the behavior is lost when $\V{M}$ alone takes its alternative values. The pure indirect effect (PIE) \citep{vanderweele2013ThreewayDecomposition} asks how much comes back when $\V{M}$ alone is restored to its factual values:\footnote{Taking $a$ as the treated level and $a'$ as the control, our $\pie$ is the pure indirect effect of \citet{vanderweele2013ThreewayDecomposition}, which equals the NIE of \citet{pearl2001direct}, and our $\nie$ is the total indirect effect of \citet{vanderweele2013ThreewayDecomposition}, which holds the treatment at its active level. We keep the names NIE and PIE to distinguish the two patching directions.}
\begin{align}
    \nie_{\V{M}}(a, a') &= Y(a, \V{M}(a)) - Y(a, \V{M}(a')), \\
    \pie_{\V{M}}(a, a') &= Y(a', \V{M}(a)) - Y(a', \V{M}(a')).
\end{align}
Here $Y(a, \V{M}(a'))$ is the model's output on $a$ with $\V{M}$ patched to the values it takes on $a'$. The NIE measures necessity and the PIE sufficiency. Interpretability calls them the \textit{noising} and \textit{denoising} directions of activation patching \citep{heimersheim2024patching}, and we average both over a task's prompt pairs. Either way the whole set gets a single number, which cannot tell us which members carry the effect or how they depend on one another. That gap matters most when components compensate for each other or cancel out \citep{wang2023interpretability}.

\paragraph{Witnesses}
\textit{Witnesses} are components downstream of $\V{M}$ that we pin to their values on the starting prompt ($a$ for the NIE, $a'$ for the PIE), so that they cannot compensate for an intervention on $\V{M}$. In the rock throwing example, if we hold fixed that Billy's rock never hits, the bottle now survives whenever Suzy does not throw, and her throw regains its effect. \Cref{fig:estimand-sets} shows the same result in the IOI circuit. Scored alone, the backup name movers outrank the primaries (NIE $0.74$ against $0.14$); with witnesses integrated, the primaries lead ($1.16$ against $0.60$ for the backups).

With witnesses $\V{W}$, the \textit{witness-informed} NIE and PIE are:
\begin{align}
\label{eq:nie-witness}
\nie_{\V{M}; \V{W}}(a, a') &\;\triangleq\; Y(a, \V{M}(a), \V{W}(a)) - Y(a, \V{M}(a'), \V{W}(a)), \\
\pie_{\V{M}; \V{W}}(a, a') &\;\triangleq\; Y(a', \V{M}(a), \V{W}(a')) - Y(a', \V{M}(a'), \V{W}(a')). \label{eq:pie_witness}
\end{align}
Each witness is pinned to the value it takes on the prompt the run starts from, which is $a$ for the NIE and $a'$ for the PIE.

\paragraph{Witness Integration}
Since we rarely know in advance which witnesses to pin, we average over a distribution of witness sets \citep{urbaniak2026pci}. Let $\V{V}$ be all the variables of the model and $\V{C} \subseteq \V{V}$ the pool of \textit{candidates} we search over. Each draw splits $\V{V}$ three ways. The mediators $\V{M} \subseteq \V{C}$ get patched. The witnesses $\V{W} \subseteq \mathrm{reach}(\V{M})$ get pinned, where $\mathrm{reach}(\V{M})$ is everything downstream of $\V{M}$ outside $\V{M}$ itself. The remaining free variables $\V{F}$ recompute as usual. For a distribution $p(\V{M}, \V{W} \given \V{C})$ over mediators and witnesses, we define the \emph{witness-integrated set effect (WISE)} of any set-level score $\mathcal{S}$ as
\begin{equation}
    \WISE_{p,\mathcal{S}}(a, a') \;\triangleq\; \E_{p(\V{M}, \V{W} \given \V{C})}\bigl[\mathcal{S}_{\V{M}; \V{W}}(a, a')\bigr].
\end{equation}
Setting $\mathcal{S} = \nie$ gives the \emph{witness-integrated NIE}, $\WINIE_p \triangleq \WISE_{p, \nie}$, the objective that our \textsc{JuntaLearner} usually maximizes (\Cref{sec:methods}), and $\mathcal{S} = \pie$ gives $\WIPIE_p \triangleq \WISE_{p, \pie}$.
\footnote{Per-component scores can be read off WISE by contrasting sampled sets with and without a component; see \Cref{app:sampling} for details.}

Sets and witnesses expose \textit{interaction effects}, the part of a joint effect that per-head evaluation typically misses. When a primary component and its backup fall in the same mediator set, patching them together exposes their joint effect; when the backup is left outside the mediator set, pinning it as a witness does the same. Set sampling and witness pinning are partly overlapping ways of exposing interactions, and how much witnesses add depends on the model; in our experiments it helps most on Qwen3. See \Cref{sec:results} and \Cref{fig:knobs,fig:knobs-bytask} in \Cref{app:overall} for details.

Enumerating mediator and witness sets becomes impractical as the candidate pool grows. Monte Carlo sampling over set sizes and witnesses can approximate WISE; details are in\Cref{app:sampling}. However, every candidate needs many sampled sets, each of which gets a forward pass, so it scales poorly to large models. In the next section, we describe a gradient-based learner that uses  $\WINIE$ maximization to identify circuits more efficiently.

\section{Circuit Discovery with \textsc{JuntaLearner}} \label{sec:methods}

Since scoring every set of candidates is out of reach, \textsc{JuntaLearner} learns which sets are worth scoring. Each candidate has a trainable inclusion logit. At each step, the method draws sets from these logits, scores each set on paired prompts, and updates the logits to favor sets with higher scores. The set size starts large and shrinks as training proceeds. At the end, a readout of the run ranks the candidates, and the top $\ell$ form the circuit of size $\ell$. \Cref{app:learning} gives the full procedure.

\paragraph{Objective}
Let $\theta_i$ be the inclusion logit of candidate $C_i \in \V{C}$. At set size $k$, the logits and the witness-sampling rule define a distribution $p_{\theta,k}(\V{M},\V{W}\given\V{C})$ over mediator--witness pairs with $|\V{M}|=k$. Training maximizes the witness-integrated score of \Cref{sec:estimand} under that distribution:
\begin{equation}
    \max_{\theta}\;
    \WISE_{p_{\theta,k},\mathcal{S}}(a,a')
    =
    \max_{\theta}\;
    \E_{p_{\theta,k}}
    \bigl[\mathcal{S}_{\V{M};\V{W}}(a,a')\bigr].
    \label{eq:7}
\end{equation}
With $\mathcal{S}=\nie$ this is $\WINIE$, taken over sets of size $k$. The choice of $\mathcal{S}$ decides what the ranking looks for: $\nie$ trains toward necessity, and $\pie$ toward sufficiency. We may also obtain a hybrid score by taking the soft minimum of $\nie$ and $\pie$, which we call the \emph{bottleneck score}:\footnote{As $\tau\to 0$ it approaches $\min(\nie,\pie)$; for large $\tau$ it behaves like their mean shifted by $-\tau\log 2$ and tends to $-\infty$, so $\tau$ should be small relative to the scale of the effects.}
\begin{equation}
\hat{\mathcal{S}}_{\V{M};\V{W}}(a,a')
\triangleq
-\tau\log\!\left(
e^{-\nie_{\V{M};\V{W}}(a,a')/\tau}
+
e^{-\pie_{\V{M};\V{W}}(a,a')/\tau}
\right),
\qquad \tau>0, \label{eq:bottleneck}
\end{equation}
A candidate's gradient tracks how the expected score changes as that candidate becomes more likely to be drawn.

\paragraph{Sampling and updates}
Each step takes $P$ prompt pairs and draws $B$ mediator sets from the logits. A set is drawn by adding independent Gumbel noise to every logit and keeping the $k$ largest:
\begin{equation}
    g_i \overset{\mathrm{iid}}{\sim} \mathrm{Gumbel}(0,1),
    \qquad
    \V{M} = \{C_i : i \in \operatorname{top}_k(\theta + g)\}.
\end{equation}
For each mediator set, we sample witnesses uniformly from its eligible downstream candidates, with the witness count determined by a chosen fraction $m$ of that eligible pool. With $m=0$ no witnesses are pinned, and the score is the plain set-level NIE or PIE. All $B$ sets are scored on the same prompt pairs. The forward pass uses the hard selection, so every component is either patched, pinned or left free to recompute. The backward pass uses a straight-through estimator \citep{jang2017categorical}, which takes its gradient from the continuous Gumbel-top-$k$ relaxation of \citet{Xie2019Reparameterizable}. The Adam optimizer with gradient clipping then updates the logits.

\paragraph{Shrinking the set size}
Training follows a descending ladder of set sizes $k_1 > \cdots > k_R$, carrying the logits and the optimizer state from one rung to the next. Large sets allow training to find effects that appear only when several components are patched together, and the smaller sets that follow help separate the members of a good set. A rung ends early once its top candidates have stopped changing from step to step, without flipping back and forth, and the logit gap at the $k_r$-th place is still widening; see \Cref{def:earned} for details. Otherwise it ends after $\lfloor T/R\rfloor$ steps of the total budget $T$. The last rung runs until the budget is spent, so it also gets the steps that earlier rungs saved. In our experiments, training over decreasing set sizes improves all evaluation scores on every model compared with training only on single-component sets throughout (\Cref{sec:results}).

\paragraph{Ranking}
Since Adam adjusts the update scale separately for each inclusion logit, final logits need not preserve the relative magnitudes of candidates' effects and hence would make a poor ranking.
We therefore rank candidates using gradients recorded during training, normalized per step and clipped per candidate (\Cref{eq:readout}). The normalization and clipping are engineering choices
 that are not dictated by the estimands in \Cref{sec:estimand}; we describe them and how we set them in \Cref{app:learning}. Writing $C_{(1)},\ldots,C_{(n)}$ for the ranked candidates, the circuit of size $\ell$ is $\{\mathcal{C}_{(1)},\ldots,\mathcal{C}_{(\ell)}\}$, and one training run yields the whole nested family of circuits. \Cref{alg:junta} summarizes the procedure, with $\widehat{\mathcal{S}}$ for the score averaged over the step's prompt pairs and $\widetilde{\nabla}_{\theta}$ for the straight-through gradient.

\begin{algorithm}[t]
\caption{\textsc{JuntaLearner}}
\label{alg:junta}
\begin{algorithmic}[1]
\Require Candidate pool $\V{C}$; score $\mathcal{S}$;
ladder $k_1,\ldots,k_R$; step budget $T$;
$B$ sets and $P$ prompt pairs per step; witness fraction $m\in[0,1]$
\Ensure $\mathrm{score}(i)$ for every candidate $C_i\in\V{C}$
\State $\theta\gets\mathbf{0}$ \Comment{one inclusion logit per candidate}
\For{$r=1,\ldots,R$}
    \While{budget remains and ($r=R$ or rung $r$ has not ended)}
        \State Take $P$ prompt pairs without replacement
        \For{$b=1,\ldots,B$}
            \State $g\sim\mathrm{Gumbel}(0,1)^{|\V{C}|}$
            \State $\V{M}_b\gets \{C_i : i\in\operatorname{top}_{k_r}(\theta+g)\}$
            \Comment{straight-through gradients}
            \State $\mathcal{E}_b\gets$ components downstream of $\V{M}_b$
            \State $\V{W}_b\sim\mathrm{Unif}\bigl(
            \{W\subseteq\mathcal{E}_b:
            |W|=\lfloor m|\mathcal{E}_b|\rceil\}\bigr)$
        \EndFor
        \State $\gamma \gets -\widetilde{\nabla}_{\theta}
        \frac{1}{B}\sum_{b=1}^{B}
        \widehat{\mathcal{S}}_{\V{M}_b;\V{W}_b}$; record $\gamma$
        \State $\theta\gets\mathrm{Adam}(\theta,\gamma)$
        \Comment{with gradient clipping}
    \EndWhile
\EndFor
\State $\mathrm{score}(i)\gets\sum_t \operatorname{clip}(\hat\gamma_{t,i})$, over the recorded $\gamma_t$ normalized per step
\Comment{\Cref{eq:readout}}
\end{algorithmic}
\end{algorithm}

\section{Experiments} \label{sec:results}

We test how well \textsc{JuntaLearner} can localize task behavior to compact circuits across models and tasks. We evaluate on GPT-2 small, Qwen3-1.7B-Base, and Llama-3.1-8B, using their attention heads as candidate causes and witnesses. The outcome $Y$ is the correct-answer logit minus the incorrect-answer logit at the final position. For each model, we study performance on indirect object identification (IOI; \citealp{wang2023interpretability}), subject--verb agreement \citep{finlayson2021causal}, two-digit addition \citep{stolfo2023mechanistic}, induction \citep{elhage2021mathematical, olsson2022induction}, and entity tracking \citep{prakash2023fine}.
We provide experimental settings and examples of each task in \cref{app:experiments}.

We compare \textsc{JuntaLearner} against rankings obtained by single-component activation patching (per-head NIE), EAP \citep{syed2024attribution}, and EAP-IG \citep{hanna2024have}, evaluating rankings by both signed scores and absolute score magnitudes. We also compare against Matryoshka Attribution (MAttr; \citealp{arora2026matryoshka}) and random rankings. The MIB Circuit Localization track \citep{mueller2025mib} gets a separate comparison in \cref{app:mib}.

\subsection{Circuit Evaluation}

Recovering the model's task performance is the most well-known dimension of circuit quality. \emph{Faithfulness} measures this by ablating components outside the circuit and testing how much performance remains \citep{wang2023interpretability}, and faithfulness across circuit sizes is often summarized using aggregate metrics such as CPR and CMD \citep{mueller2025mib}. While faithfulness establishes how much performance the circuit can support, it does not tell us how much the intact model depends on it: components elsewhere may compensate when the circuit is removed. Moreover, a large removal effect may reflect damage to general-purpose components that support many tasks. We also ask how few components are needed to achieve these effects, to reflect the goal of localizing behavior to small circuits. We therefore supplement standard circuit metrics with measures of removal effects and task specificity, and aggregates across circuit sizes that give greater weight to small circuits.

\paragraph{Faithfulness and Necessity}
Let the circuit $\mathcal{C}_{\ell}$ contain a ranking's top $\ell$ candidates, and write $Y_{\mathcal{C}_{\ell}}= Y(a,\mathcal{C}_{\ell}(a'))$ for the outcome when components in $\mathcal{C}_{\ell}$ are patched to their alternative-prompt values. Faithfulness measures the performance retained when the circuit's complement is patched, while \emph{kill} measures necessity, the effect of patching the circuit itself. In both cases, the remaining components recompute normally, with no witnesses pinned. Then:

\[
\operatorname{Faithfulness}(\ell)
= \frac{\E[Y_{\V{C}\setminus\mathcal{C}_{\ell}}]-\E[Y_{\V{C}}]}
       {\E[Y_{\varnothing}]-\E[Y_{\V{C}}]},
\qquad
\operatorname{Kill}(\ell)
= \frac{\E[Y_{\varnothing}]-\E[Y_{\mathcal{C}_{\ell}}]}
       {\E[Y_{\varnothing}]-\E[Y_{\V{C}}]}.
\]

\begin{wrapfigure}[20]{l}{0.34\textwidth}
\vspace{-12pt}
\centering
\includegraphics[width=0.34\textwidth]{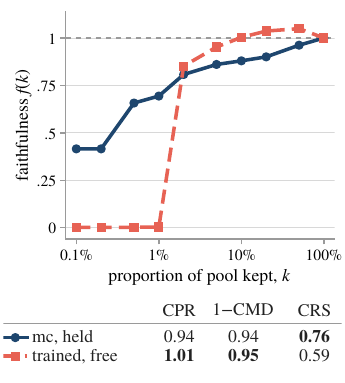}
\caption{Two faithfulness curves on Qwen3-1.7B-Base, SVA, at MIB's ten log-spaced proportions. CPR and $1-\mathrm{CMD}$ prefer the late, overshooting curve; CRS prefers the early one.}
\label{fig:early-recognition}
\vspace{-10pt}
\end{wrapfigure}

All expectations are over held-out prompt pairs, and both metrics are normalized by the effect of patching the full candidate pool. The numerator of kill is the circuit's NIE. Faithfulness reaches 1 when the circuit recovers the intact model's mean outcome; kill reaches 1 when removing the circuit has the same effect as patching the full pool.

\paragraph{Task Specificity}
To measure task specificity, we apply the same circuit $\mathcal{C}_{\ell}$
to the set $\mathcal{T}$ of the model's other tasks.\footnote{All four others on Qwen3 and Llama, and on GPT-2 the others except entity tracking, which GPT-2 small barely performs.} Let $\operatorname{Kill}_j(\ell)$ denote the removal effect on task $j$ using its own prompt pairs and normalization. We define \emph{net} as target-task removal minus mean collateral damage:
\[
\operatorname{Net}(\ell)
=
\min\{1,\operatorname{Kill}(\ell)\}
-
\frac{1}{|\mathcal{T}|}
\sum_{j\in\mathcal{T}}
\max\{0,\operatorname{Kill}_j(\ell)\}.
\]
We cap the target effect at one so that removing the circuit earns no extra credit for causing more damage than patching all candidate components. We floor each control effect at zero so that improving one task cannot offset damage to another. Higher net values indicate greater target-task disruption relative to damage on the other tasks.

\paragraph{Aggregation Across Circuit Sizes}
A ranking localizes behavior more compactly when small circuits already achieve the desired effect. Faithfulness is usually summarized using the integrated circuit performance ratio (CPR), the area under the faithfulness curve, and the integrated circuit-model distance (CMD), the area between that curve and one \citep{mueller2025mib}. CPR rewards greater recovered performance, including performance that overshoots the intact model's, whereas CMD penalizes deviations in either direction (lower CMD is better; in figures we plot $1-\mathrm{CMD}$, so that higher is better on every summary). Both integrate over circuit size on a linear scale, so performance at circuit sizes of 1--2\% of the candidate pool receives the same weight as performance at sizes of 50--51\%, where nearly every ranking has already recovered the behavior.

To prioritize small circuits, we adopt the principle of rewarding early success used in discounted cumulative gain for information retrieval \citep{jarvelin2002cumulated} and early-recognition metrics for virtual screening \citep{truchon2007bedroc}, and average performance over circuit sizes on a logarithmic scale. For each evaluation curve $h$, we define its \emph{circuit recognition score} (CRS) over evaluated sizes $\ell_1<\cdots<\ell_J$, with $\ell_1>0$, as
\[
\operatorname{CRS}(h)
=
\frac{\sum_{j=1}^{J} w_j\,\min\{1,h(\ell_j)\}}
     {\sum_{j=1}^{J} w_j}.
\]
The weights $w_j$ are the local spacings of $\log\ell_j$, computed by central differences at interior points and one-sided differences at the endpoints. Faithfulness is evaluated at MIB's ten log-spaced proportions of the candidate pool, and kill and net at the top $\ell=1,\ldots,25$ heads (26 on some cells). Capping faithfulness and kill at one gives no additional credit or penalty for exceeding their reference outcomes; net is already bounded above by one by construction. Applying CRS to the three curves gives $S_{\mathrm{faith}}$, $S_{\mathrm{kill}}$, and $S_{\mathrm{net}}$, respectively.

\Cref{fig:early-recognition} shows how the choice of summary can reverse a verdict. One curve recovers the behavior from the smallest circuits onward; the other stays at zero until about 1\% of the pool and then overshoots. CPR and $1-\mathrm{CMD}$ rank the late curve higher (1.01 and 0.95 against 0.94 and 0.94), and CRS ranks the early one higher (0.76 against 0.59). We report CRS in the main text; \Cref{app:mib} reports all three summaries side by side.

\subsection{Results}

\begin{figure}[t]
\centering
\includegraphics[width=\linewidth]{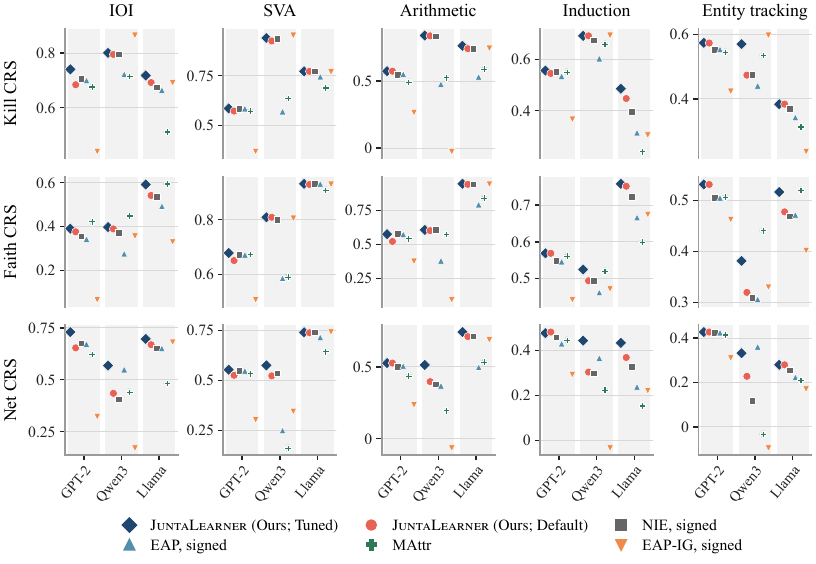}
\caption{Every (model, task) cell, scored by CRS. \textit{Per-model default} is one
configuration per model, chosen by median per-cell score; \textit{Tuned} is, for each cell and each goal, the \textsc{JuntaLearner} configuration with the highest CRS on that goal (mean over seeds)}
\label{fig:robustness}
\end{figure}

\begin{table}[t]
\centering
\small
\setlength{\tabcolsep}{3.5pt}
\begin{tabular}{lrrrrrr}
\toprule
 & \multicolumn{2}{c}{Kill ($\uparrow$)} & \multicolumn{2}{c}{Faithfulness ($\uparrow$)} & \multicolumn{2}{c}{Net ($\uparrow$)} \\
Method & Mean & Worst& Mean & Worst & Mean  & Worst \\
\midrule
\textsc{JuntaLearner} (Ours; Tuned) & \textbf{0.67} & \textbf{0.38} & \textbf{0.61} & \underline{0.38} & \textbf{0.54} & \textbf{0.28} \\
\textsc{JuntaLearner} (Ours; Default) & \underline{0.65} & \textbf{0.38} & \underline{0.59} & 0.32 & \underline{0.48} & 0.21 \\
Per-head NIE, signed & 0.64 & \underline{0.37} & 0.59 & 0.31 & 0.47 & 0.12 \\
EAP, signed & 0.56 & 0.31 & 0.53 & 0.28 & 0.45 & \underline{0.22} \\
MAttr & 0.55 & 0.24 & 0.58 & \textbf{0.42} & 0.36 & -0.03 \\
Per-head NIE & 0.56 & 0.29 & 0.50 & 0.22 & 0.40 & 0.13 \\
EAP-IG, signed & 0.51 & -0.03 & 0.48 & 0.07 & 0.29 & -0.10 \\
EAP & 0.39 & -0.07 & 0.43 & 0.15 & 0.31 & -0.11 \\
EAP-IG & 0.44 & 0.13 & 0.44 & 0.24 & 0.20 & -0.11 \\
\bottomrule
\end{tabular}
\caption{CRS across the fifteen model-task cells. Worst is the lowest score across all cells; higher is better in every column. Best results are bold and second-best results are underlined, ranked before rounding. Baselines marked "signed" order heads by signed score instead of magnitude. \cref{fig:robustness} shows the per-cell variation in scores.}
\label{tab:robustness-summary}
\end{table}

\begin{figure}[t]
\centering
\includegraphics[width=0.86\linewidth]{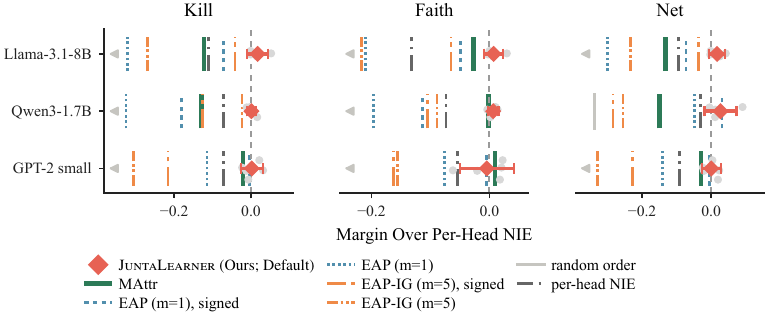}
\caption{Each model's default against signed per-head $\NIE$, pooled over its five
tasks, with a $95\%$ interval across tasks. Vertical rules mark the other baselines'
margins on the same axis. Arrows denote that random order is lies significantly off-axes, and would distort the scale to be included.}
\label{fig:uncertainty}
\end{figure}

We evaluate two configuration-selection strategies across the fifteen model--task combinations. \emph{Default} uses one configuration per model, selected by its median score across the five tasks. \emph{Tuned} uses task-specific configurations selected from the sweep (\Cref{app:experiments}). We compare the resulting rankings on faithfulness, kill, and net, summarized by CRS.

\textbf{JuntaLearner outperforms attribution baselines on average.}
The tuned configurations achieve the highest mean faithfulness, kill, and net scores, leading five of six columns in \Cref{tab:robustness-summary}. With one fixed configuration per model, \textsc{JuntaLearner} scores comparably to signed exact per-head NIE and higher than EAP, EAP-IG, and MAttr. Worst-cell net scores separate the methods further: MAttr, both EAP-IG variants, and magnitude-ranked EAP each have at least one negative-net cell. \Cref{fig:robustness} shows per-cell scores: on GPT-2 small, our default, signed per-head NIE, and signed EAP are within a few hundredths on nearly every panel; on Llama-3.1-8B, \textsc{JuntaLearner} default exceeds all attribution baselines on each metric when pooled across tasks. \Cref{fig:uncertainty} reports margins over signed per-head NIE with $95\%$ intervals across seeds.

\paragraph{The baselines' strengths are metric-specific.}
On Qwen3-1.7B, signed EAP-IG matches the strongest methods on kill for four of five tasks, yet its net score is negative on three---its top-ranked heads recur across tasks (mean IoU $0.28$ vs.\ $0.12$ for exact per-head NIE). MAttr shows the opposite profile: highest worst-cell faithfulness ($0.42$), beating our default on faithfulness in eight of fifteen cells, while our default exceeds it on kill in twelve and on net in thirteen. These differences reflect training objectives: MAttr patches the complement of the circuit, which is what faithfulness measures; \textsc{JuntaLearner} patches the circuit itself, which is what kill measures.

\begin{wrapfigure}{r}{0.44\textwidth}
\vspace{-0.7\baselineskip}
\centering
\includegraphics[width=\linewidth]{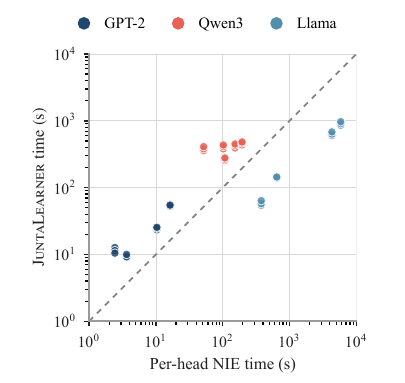}
\caption{Training time for each model--task--regime run (seconds, log scales).
The dashed line marks equal time; points below it favor \textsc{JuntaLearner}.}
\label{fig:timing}
\vspace{-0.7\baselineskip}
\end{wrapfigure}

\paragraph{Training is faster than exact scoring on the largest model.}
On Llama-3.1-8B, \textsc{JuntaLearner} trains in $14$--$16$ minutes on the slowest task, versus $97$ minutes for exact per-head NIE---a $4$--$7\times$ speedup that holds across all timed runs (\Cref{fig:timing}). On the smaller models, training is $2$--$8\times$ slower than exact patching, which takes seconds to minutes. The difference is structural: exact scoring costs $\Theta(|M| \cdot N)$ forward passes, scaling with the candidate pool; training costs $\Theta(T \cdot B \cdot P)$, independent of pool size.

\paragraph{JuntaLearner tuning favors descending set sizes and necessity objectives, with model-specific gains from witness pinning.}

Each run is shaped by five knobs: whether to descend a ladder of set sizes or train on single candidates throughout, how wide to start the ladder (as a share of the candidate pool), which objective to score ($\NIE$, $\NIE+\pie$, or the bottleneck score of \cref{eq:bottleneck}), how many downstream witnesses to pin (none, half, or all), and whether to add a final measurement stage that scores leading candidates individually. We fit a Gaussian process to each model's training runs and select settings by expected improvement, double-checked by CatBoost (\cref{fig:knobs}). Compared with training on single components throughout, the descending ladder improves predicted faithfulness, kill, and net on all three models. Starting the ladder with larger sets further improves kill and net, with a small loss in faithfulness, since large early sets favor heads whose removal matters in combination. Scoring sampled sets for necessity alone performs at least as well as combining necessity and sufficiency through either their sum or the bottleneck score. The optional measurement stage has a smaller effect, changing no model--metric comparison by more than $0.4$ posterior standard deviations.

\FloatBarrier

The benefit of witness pinning depends on the model and metric. Pinning all eligible witnesses lowers all three scores on GPT-2 and Llama. On Qwen3, it improves net while reducing kill and faithfulness, with normalized changes of $+0.7$, $-1.8$, and $-2.5$, respectively. This tradeoff also appears in the task-specific results: configurations with witnesses pinned achieve the highest net scores on all five tasks in Qwen3. This is potentially suggestive of many multi-functional backup heads in Qwen3, though we do not further investigate this claim in this work.

\paragraph{Evaluation on MIB}

We extend the comparison to the MIB Circuit Localization track \citep{mueller2025mib}, covering twelve model--task combinations across four models and six tasks. We compare \textsc{JuntaLearner} with exact per-node NIE and MAttr's released circuits on the public test split, with one shot untuned runs. Faithfulness uses MIB's native evaluator over circuits containing attention heads, MLPs, and the input embedding; we additionally evaluate kill and net over the attention heads in each ranking. We selected necessity as the score for training sets using GPT-2 IOI validation data; further protocol details and sensitivity analyses appear in \Cref{app:mib}.

Across the twelve cells, \textsc{JuntaLearner} achieves better aggregate kill and net scores than MAttr under all three summaries, and exceeds exact per-node NIE on CRS for faithfulness, kill, and net. MAttr's mean faithfulness curve rises earlier, giving it a higher CRS, and its higher CPR comes mostly from overshoot (\Cref{fig:mib-overshoot-mib}). \textsc{JuntaLearner}'s circuits do far better on CMD since it penalizes overshoot: summed over the twelve cells, their $1-\mathrm{CMD}$ is $9.1$ against MAttr's $1.0$ (\Cref{fig:mib-triangle}).

\section{Discussion}
\label{sec:discussion}

In the presence of interactions, recovering a behavior and selectively disabling it need not lead to the same circuit. A set of components can support the behavior while backups elsewhere preserve it after removal; removing a different set can disrupt the behavior through damage to machinery shared across tasks. WISE and \textsc{JuntaLearner} account for how components contribute in the context of other sets of components. The clearest sign that component sets matter comes from training itself: the descending ladder improves scores over singleton training on all metrics. Additionally, \textsc{JuntaLearner}'s higher net scores indicate circuits that are more localized to the specific task being studied, which helps with controlling its behavior in isolation.

However, the choice of evaluation metric can reverse a verdict on circuit quality. Strong removal effects alone do not establish task specificity; several baselines achieve competitive kill scores yet negative net scores. MAttr matches or exceeds the \textsc{JuntaLearner} default on faithfulness in about half of cases, particularly with CPR, but much of its lead comes from overshoot and CMD penalizes this. Our default is also ahead on kill and net in the majority of cases. These findings motivate reporting necessity and task specificity alongside faithfulness, with multiple summaries of each curve.

Several limitations remain. Our main experiments focus only on attention heads, and on MIB we ran a single untuned configuration. Training is slower than exact per-head patching on the smaller GPT-2 and Qwen3 models, and the cost advantage appears only once the pool is large. The benefit of witness pinning depends on the model and was primarily seen on Qwen3; a principled account of this dependence would clarify when and why witnesses will help identify context-dependent mechanisms.

\subsection*{Contribution Statement}

SV and RU jointly led the project. SV, RU, and EB developed and implemented the core samplers, forward methods, and dataset formatting. SV proposed using witness-based methods for circuit discovery, and SV and RU jointly proposed its initial mathematical formulation. RU designed and implemented the \textsc{JuntaLearner} training algorithm, suggested new metrics, and ran the experiments. DW and MN contributed to the WISE estimand and its supporting theory. SV, RU, and DW led the writing of the paper.

\subsection*{AI Use Statement}

Human authors are solely responsible for the research idea, theoretical analysis, interpretation of results, experiment design, and final editing and writing of this research. Generative AI (specifically, Anthropic's Claude Sonnet 5 and Opus 5, through the Claude Code CLI, and GPT-6 Astra through the Codex CLI) were used in the engineering and experimental work behind this paper in order to re-format code, debug code, and execute the experimental sweeps. The authors approved all runs in the experimental sweeps. We have reviewed all AI-assisted work. Every number and every figure in the paper is regenerated by a script in the codebase from saved run artifacts. We take responsibility for the final content of this work, including artifacts produced with the aid of generative AI.

\clearpage
\bibliography{references_iclr}
\bibliographystyle{iclr2027_conference}

\clearpage
\appendix
\crefalias{section}{appendix}

\section{Sampling algorithm} \label{app:sampling}

This paper builds a per-candidate ranking in two ways. The sampled ranking described
here is based on a Monte Carlo estimate of the $\WISE$
score of \Cref{sec:estimand} and some similar scores: it draws candidate sets, scores each one as a set, and
reads a candidate's contribution off the difference between the sets that hold it and
the sets that do not. The trained ranking of \Cref{app:learning}, roughly speaking,
learns an inclusion distribution and is read off the training trace.
Everything in \Cref{sec:results} uses the trained ranking, as MC is costly for larger
models like Llama-3.1-8B. We do use the MC estimands on smaller models to look at the
overlaps between the resulting rankings (\Cref{app:learning}) for the family of
estimands introduced in \Cref{sec:estimand}.

For simplicity we describe the sampler with witnessed NIE as the objective, which can
be swapped for any of the others.

\begin{algorithm}[H]
\caption{Sampling. One pass over the sweep, then one score per candidate.}
\label{alg:sampling}
\begin{algorithmic}[1]
\Require pool $\mathcal{P}$; sizes $\mathcal{K}$ with a draw budget $n_k$ for each; a witness regime
\Ensure $\mathrm{score}(i)$ for every candidate $i \in \mathcal{P}$
\State $\mathcal{D} \gets \emptyset$ \Comment{scored sets}
\For{$k \in \mathcal{K}$ with $k < |\mathcal{P}|$} \Comment{$k = |\mathcal{P}|$ splits nothing}
  \For{$n_k$ draws}
    \State $S \sim \mathrm{Unif}\bigl(\{\,S \subseteq \mathcal{P} : |S| = k\,\}\bigr)$
    \State $W \gets$ the regime's witness draw on $\mathrm{reach}(S)$
    \State $\mathcal{D} \gets \mathcal{D} \cup \{(S, \NIE_{S; W})\}$
  \EndFor
\EndFor
\For{$i \in \mathcal{P}$}
  \State $\mathcal{K}_i \gets \{\,k \in \mathcal{K}$ : some draw at size $k$ holds $i$ and some does not$\,\}$
  \State $\mathrm{score}(i) \gets \dfrac{1}{|\mathcal{K}_i|}\displaystyle\sum_{k \in \mathcal{K}_i}
     \Bigl( \operatorname*{mean}_{\substack{(S,v) \in \mathcal{D} \\ i \in S,\ |S| = k}} v
          - \operatorname*{mean}_{\substack{(S,v) \in \mathcal{D} \\ i \notin S,\ |S| = k}} v \Bigr)$
\EndFor
\end{algorithmic}
\end{algorithm}

The key element in the above is the matching of sets within a size. A larger set
removes more of the margin whoever is in it, so a candidate drawn mostly into large sets
would otherwise be credited for the size. The price is that the number
is a contrast a candidate with no causal role is not
guaranteed a score of zero (only the induced ordering is every used to generate circuits, though).

\section{Learning algorithm} \label{app:learning}

Most of a model's heads do nothing for any given task, and the learner looks for the few
that do. It trains one inclusion logit per candidate, saying how likely that candidate is
to be switched on, in order to maximize the causal effect. Training then repeats the
following steps: draw sets of candidates from those logits (and witnesses, if any), run
forward computing the causal effect, and backpropagate to move the logits.

Candidate set size is annealed. Typically, training starts with some set size $k_1$
broad enough to quickly explore a larger number of candidates at a high level, and ends
with $k_R$, narrow enough that a set's score says something about single candidates. The
strategy is that of the twelve-ball balance puzzle, where three weighings find the odd
ball only because each one puts groups of decreasing size on the pans. Here $k$ falls for
analogous reasons.
We call the sequence of sizes a ladder and one size a rung. A run descends the ladder and
stays on each rung until that rung is earned or its step budget runs out.

\begin{algorithm}[t]
\caption{Learning. One pass down the cardinality ladder.}
\label{alg:learning}
\begin{algorithmic}[1]
\Require pool $\mathcal{P}$; effect score $\mathcal{S}$; ladder $k_1, \dots, k_R$ in the order the schedule visits them; step budget $T$; $B$ sets and $P$ prompt pairs per step; witness fraction $m\in[0,1]$
\Ensure $\mathrm{score}(i)$ for every candidate $i \in \mathcal{P}$
\State $\theta \gets \mathbf{0}$ \Comment{one inclusion logit per candidate}
\State $C \gets \lfloor T/R \rfloor$ \Comment{each rung's step ceiling}
\For{$r = 1, \dots, R$}
  \While{rung $r$ is neither earned nor at its ceiling} \Comment{rung $R$ runs until $T$ is spent}
    \State take $P$ prompt pairs without replacement
    \For{$b = 1, \dots, B$}
      \State $g \sim \mathrm{Gumbel}(0, 1)^{|\mathcal{P}|}$
      \State $S_b \gets \mathrm{top}_{k_r}(\theta + g)$ \Comment{straight-through, so $\partial S_b / \partial \theta$ exists}
      \State $W_b \sim \mathrm{Unif}\bigl(\{\,W \subseteq \mathrm{reach}(S_b) : |W| = \lfloor m\,|\mathrm{reach}(S_b)|\rceil\,\}\bigr)$
    \EndFor
    \State record $\nabla_\theta$ for this step
    \State $\theta \gets \mathrm{Adam}\Bigl(\theta,\ -\nabla_\theta \tfrac{1}{B} \textstyle\sum_{b} \mathcal{S}_{S_b; W_b}\Bigr)$
  \EndWhile
\EndFor
\State $\mathrm{score}(i) \gets$ the readout of \Cref{eq:readout} over the recorded gradients
\end{algorithmic}
\end{algorithm}

One step of \Cref{alg:learning} goes as follows. It takes $P$ prompt pairs and draws $B$
candidate sets of the rung's size $k_r$, each set with its own witness subset drawn from
what that set reaches. It scores every set with the witness-integrated objective
$\mathcal{S}$ on those pairs. It then takes one Adam step on the logits. The draw is a
top-$k$ Gumbel sample with a straight-through estimator, which makes a hard set of $k_r$
candidates differentiable in $\theta$. Prompt pairs are taken without replacement from a
shuffled pass over the training split, refilled when the pass runs out, so no pair is
seen much more often than another. And all $B$ sets in a step score the same pairs.

\paragraph{Advancing a rung.} A rung ends one of two ways. Either the logits settle, and
the rung is \emph{earned}, or the rung runs out of steps and ends \emph{at its ceiling} (the last rung always runs
until the budget is spent). Settling means three things at once: the top of the ranking stops changing from step to
step, it is not merely oscillating, and the gap at the cut is still widening.

\begin{definition}[Earned rung]
\label{def:earned}
Rung $r$ trains at cardinality $k_r$ over a sequence of steps, each ending with an Adam
update to $\theta$. At step $t$, let $w = \max(k_r, \min(10, |\mathcal{P}|))$ and
$\mathrm{supp}_t = \mathrm{top}_w(\theta_t)$, the index set of the top $w$ logits. $w$ is
wider than $k_r$ itself: at $k_r=1$ a one-element set makes overlap binary, reporting
only whether the leader changed. The step-to-step overlap is
$
o_t = \frac{|\mathrm{supp}_t \cap \mathrm{supp}_{t-1}|}{|\mathrm{supp}_t \cup \mathrm{supp}_{t-1}|},
$
and the separation margin $\mu_t$ is the gap between the $k_r$-th and $(k_r{+}1)$-th
largest entries of $\theta_t$, normalized by the spread between the median of its top
fifth and the median of the whole vector.

Let $C = \lfloor T/R \rfloor$ be the rung's step ceiling, $\alpha = \min(1, 3/C)$, and
$\mathrm{span} = \max(2, \mathrm{round}(1/\alpha))$, about a third of $C$. Let
$\bar{o}_t$ be the exponential moving average of $o_t$ with decay $\alpha$. Rung $r$ is
\emph{earned} at the first step $t \ge \mathrm{span}$ where the following three
conditions are met:\footnote{We use $\tau = 0.9$ and $\epsilon = 0.1$ throughout. Each condition rules out a different
way of being fooled. Two runs can reach the same $\bar{o}_t$: one that has settled, and
one that flips between two members every other step. Only the flipping one dips below
$\tau - \epsilon$ inside the span, and that dip is what the second condition looks for.
The third
catches a run that has gone quiet instead of resolving, where the top $k_r$ stop changing
because the logits have stopped moving at all.}
\begin{enumerate}
    \item \textbf{(Convergence)} The moving average of $o_t$ exceeds a threshold $\tau$, i.e., $\bar{o}_t \geq \tau$;
    \item \textbf{(Stability)} Over a sliding window, $o_t$ has never dropped more than $\epsilon$-far from the threshold $\tau$, defined as $\min(o_{t-\mathrm{span}+1}, \dots, o_t) \ge \tau - \epsilon$; and
    \item \textbf{(Separation)} The separation margin $\mu_t$ has improved over the beginning of the sliding window, i.e., $\mu_t \geq \mu_{t-\mathrm{span}+1}$.
\end{enumerate}
\end{definition}

\paragraph{Readout.} First of all, the optimizer in this task has quite a bit to handle. The gradient of a set
score has no fixed scale. It depends on the objective, on the model,
 on how many candidates the set holds and on which
other candidates came with them, and it falls several-fold over the course of the ladder
as $k$ narrows. Plain gradient ascent at a fixed learning rate therefore behaves
differently on every task and every rung. Adam normalization smooths things over and in a sense
removes the scale. It keeps a running estimate of each logit's own recent gradient
magnitude and divides that logit's update by it, so every logit moves about one learning
rate per step whatever its raw gradient happened to be. Only the direction and the
consistency of a candidate's gradient survive the division. That is what makes one step
budget work across models, tasks and objectives.

It is also what makes the final logit not great as a direct score. Adam normalization divides
away the effect size and so candidates can
end a run at the same logit for quite different reasons: one had a large effect on a few steps,
the other a small effect on every step. Just looking at the logits won't help you resolve this.

Note however that the gradients themselves are still in the trace. We cannot just sum
them up: with $B$ sets drawn per step, a single large gradient at a step can be an
artifact of the company a candidate happened to be drawn with, and that one step might
easily dominate the sum. So the next simple thing one could do is to (1) normalize within step
to account for changes in set sizes (and resulting changes in gradient absolute values), and  (2) cap the sum do mitigate the
impact of extreme bursts. Write $g_t \in \mathbb{R}^{|\Pset|}$ for the
negated gradient recorded at step $t$ (line 11 of \Cref{alg:learning}). Normalize each
step, then clip each candidate's contribution against its own scale:
\begin{equation}
\label{eq:readout}
\begin{gathered}
\hat g_t = \frac{g_t}{\max\bigl(\lVert g_t \rVert_2,\; 0.25\, \mathrm{median}_t \lVert g_t \rVert_2\bigr)},
\qquad
s_i = \mathrm{median}_t \lvert \hat g_{t,i} \rvert, \\[2pt]
\mathrm{score}(i) = \sum_{t} \mathrm{clip}\bigl(\hat g_{t,i},\, -c\,s_i,\; c\,s_i\bigr),
\end{gathered}
\end{equation}
with $c = 3$, and the median taken over every step, including the steps that did not draw
$i$. The per-step normalization gives every rung one unit-norm vote, because gradient
scale falls several-fold over the ladder, and an unnormalized sum
would weight the broad early rungs, where nothing is resolved yet, over the narrow late
ones where the run settles its top-$k$. The cap limits one step to $c$ times a
candidate's own typical gradient, so a burst adds at most three medians however large it
is, while a candidate with real support on most steps keeps accumulating at full
strength.

\paragraph{Measuring in readout.} \Cref{eq:readout} is an indirect score. A
candidate's gradient is read in the company of whichever other candidates its set
happened to hold, and the sum has to be normalized and capped to keep that company from
deciding the answer. Where a direct measurement of a candidate's own effect is available,
it is the quantity the ranking wants, and no inference is needed. A ladder whose last rung is $k = 1$ makes some
of those measurements for free. Every set
drawn on that rung holds exactly one candidate, so the set's score is not evidence about
that candidate's effect, it is that effect, measured on the step's prompts.
So, if the last rung actually happened to be $k=1$, one trick to test is to use
the individual effects from that stage directly to sort the final top candidates.
Training still decides who gets measured. Draws on the last rung come from the logits, so
a candidate is drawn at $k = 1$ only if the ladder has already carried it near the top.
The measurements settle the order among those finalists, assuming the signal-to-noise
ratio in this final stage is promising.\footnote{None of this applies to a ladder whose last rung
 is larger than one, since no set it draws ever holds a single candidate.
That is Llama's case: its default ladder is
$142$-$71$-$43$-$21$-$7$, so every set holds at least seven candidates, no direct
measurement is ever taken, and the ranking is \Cref{eq:readout} and nothing else.}

\paragraph{Measurement stage in training.} A final training rung at $k = 1$, as
discussed above, draws from the logits and spends most of its draws on the current
leaders, whose correct ordering matters the most for circuit discovery. While that
sounds like the right place to spend them, it isn't always the case. The reason is how the
draws are spread inside the leading group. It might happen that the candidate in front is
 drawn again and again, long after its position has stopped being in doubt, so measurement after
measurement is spent confirming something already settled. The candidates just behind it
might still be drawn once or twice each, and one or two draws cannot order them properly.
The allocation is somewhat self-reinforcing: the candidate the logits favour is the candidate
the draws keep going to.

The measurement stage conducts some measurements in a more balanced fashion to alleviate
those potential undesirable inequalities, and since we already sample them one by one, it
corrects by their own individual effects. It focuses on a set of the top
$M = \lfloor B\,T'/d \rfloor$ candidates by logit, for $B$ worlds per step, $T'$ steps
remaining and $d$ measurements per candidate, with $d = 4$ throughout. Inside that set
every candidate is measured the same number of times, the leader and the $M$-th alike,
because the stage draws them in a balanced cycle instead of drawing from the logits. Candidates outside keep the logits and gradients from the
training.

The push starts from the logit each covered candidate had when the stage began. Writing
$y_{i,m}$ for candidate $i$'s $m$-th measurement, the stage keeps a running mean:
\begin{equation}
\label{eq:measure}
\theta_i \;\leftarrow\; \theta_i + \frac{y_{i,m} - \theta_i}{m}.
\end{equation}
The first measurement has $m = 1$, so it moves $\theta_i$ the whole way onto $y_{i,1}$
and the entry logit drops out.

\begin{figure}[t]
\centering
\includegraphics[width=\linewidth]{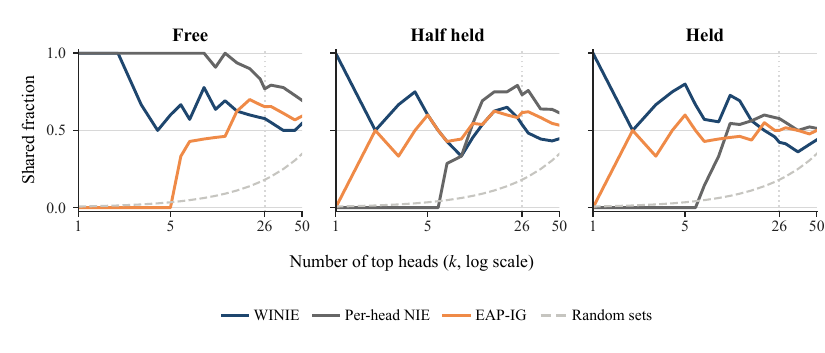}
\caption{Agreement of three score orderings with the learned \textsc{JuntaLearner} ordering
for GPT-2/IOI (seed 0). Plots the free, half-held, and held witness regimes, for the first $50$ out of $144$ attention heads.
Each curve gives the fraction of the top $k$ learned heads also in a score
ordering's top $k$. The gray dashed curve is the random-set expectation $k/|\Pset|$;
the dotted vertical line marks $k=26$. The Qwen held cell was not run.}
\label{fig:overlap}
\end{figure}

\Cref{alg:learning} optimizes an inclusion distribution while using a set-based score such as $\WINIE$,
 and never estimates the $\WISE$ score of \Cref{sec:estimand}. The relation between the resulting rankings
 and the ones that would have been established using the estimands is interesting. While deeper theoretical
 connections await exploration, we have partially answered this question empirically
(\Cref{fig:overlap}).
Each panel plots, for one (model, task) cell, the share of a reference ordering's top $k$
that three per-candidate orderings also put in their top $k$, against the $k/|\Pset|$
two random sets of that size share. Per-head $\NIE$ and $\WINIE$ score a candidate
alone, with witnesses free and integrated respectively. EAP-IG is the gradient baseline. We find that per-head NIE is most in line with \textsc{JuntaLearning} when witnesses are free, and that WINIE comes further into agreement (with modest correlation) as witnesses are held (either half-held or fully held).

Under the junta reading, the two estimands that agree most are the two exact ones at
the ends of the marginal-contribution family, LOO at the full complement and per-head
$\NIE$ at the empty context. Note also that  here agreement is not accuracy: none of the five orderings
 is ground truth. If anything is, it is the circuit performance.

\paragraph{Runtime}
Scoring every candidate set
exactly costs $\Theta(2^{|M|})$ forward passes for a pool $M$, which is infeasible. The
per-head grid replaces it with $|M|$ singleton patches, one forward pass per candidate
per prompt pair, so it costs $\Theta(|M| \cdot N)$ for $N$ pairs, linear in a pool of
$144$, $448$ or $1024$ heads here. \textsc{JuntaLearner} instead spends a fixed budget of
$T$ steps, each drawing $B$ candidate sets over $P$ prompt pairs, so it costs
$\Theta(T \cdot B \cdot P)$ forward-backward passes, independent of $|M|$. A
forward-backward pass costs a small constant multiple of a forward one, so the comparison is $|M| \cdot N$ against $T B P$. The gradient approximations cost $\Theta(m \cdot N)$ for $m$ integration steps, also independent of $|M|$, but their circuits score worse.

\section{Experiments} \label{app:experiments}

In \Cref{sec:results} we describe the models, tasks, metrics and baselines, and in \Cref{tab:tasks} we provide more details about the tasks. Here we discuss how we chose the configurations and how experimental runs were orchestrated.

\paragraph{Choosing configurations.} Each training run costs GPU time, so two surrogate models chose the $572$ configurations behind \Cref{sec:results}. After each round we fit a Gaussian process to every run so far, one per goal, with the model, the task and the knobs of \Cref{tab:knobs} as inputs and the CRS as output. We use BoTorch's \texttt{MixedSingleTaskGP}, which pairs a Mat\'ern kernel over the numeric knobs with a kernel over the categorical ones, and pick the next runs by expected improvement. Gradient-boosted trees (CatBoost), fit to the same table, serve as a check on each recommendation. The search covers $24$ to $76$ configurations per cell (\Cref{tab:runcounts}), and $150$ of the $572$ runs are seed replicates, so the per-cell best scores in \Cref{fig:robustness} show what tuning buys within a limited search.

\begin{figure}[p]
\centering
\includegraphics[width=\linewidth]{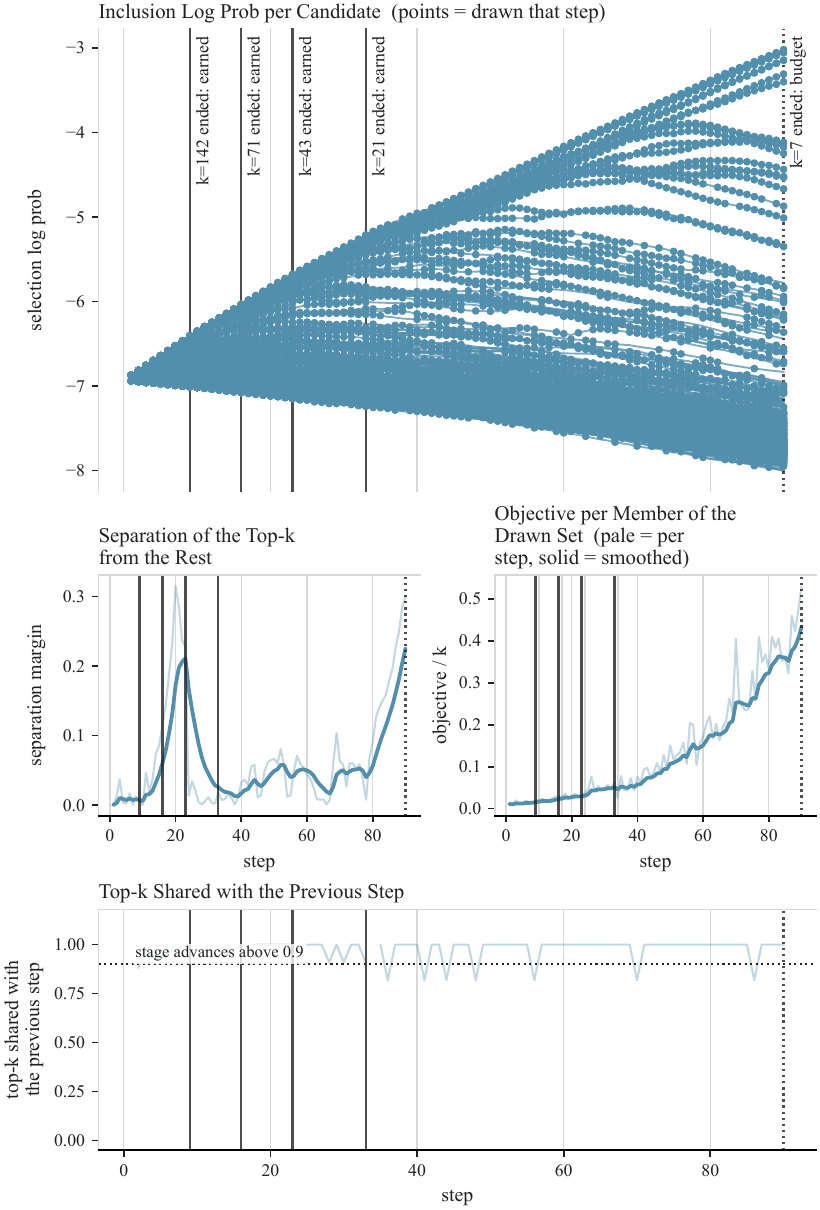}
\caption{One training run on Llama-3.1-8B: arithmetic, the free witness regime, seed
$0$, ladder $(142,71,43,21,7)$. Top: every candidate's inclusion log-prob across steps,
a point marking a step that drew it, with vertical rules where a rung ends and why.
Bottom left: the separation margin, climbing again once a new rung starts. Bottom
right: the mean objective per member of the drawn set, smoothed; a jump at a boundary
is the change in $k$. Bottom: the share of the top $k$ carried over from the previous
step, which is the quantity a rung has to hold above $0.9$ to end early.}
\label{fig:trace}
\end{figure}

\paragraph{One run.} \Cref{fig:trace} follows Llama's default on addition, the ladder $142$-$71$-$43$-$21$-$7$ with no witnesses pinned. The first four rungs each end early, once the drawn sets stop changing. The last rung runs to its step budget while the ordering is still moving, which is what it is for. The candidates' log-probabilities fan out from a flat start, and the separation between the kept candidates and the rest climbs to about $0.3$ over the final rung.

\begin{table}[t]
\centering
\caption{Distinct trained configurations by model and task.}
\label{tab:runcounts}
\small
\setlength{\tabcolsep}{7pt}
\begin{tabular}{@{}lrrr@{}}
\toprule
Task & GPT-2 small & Qwen3-1.7B & Llama-3.1-8B \\
\midrule
IOI & 51 & 32 & 44 \\
SVA & 51 & 36 & 24 \\
Arithmetic & 76 & 31 & 32 \\
Induction & 51 & 31 & 32 \\
Entity tracking & 26 & 29 & 26 \\
\midrule
Total & 255 & 159 & 158 \\
\bottomrule
\end{tabular}
\end{table}

\paragraph{Knobs and seeds.} \Cref{tab:knobs} lists the knobs we varied. Step budgets run from $30$ to $90$ steps on GPT-2 and Qwen3 and from $30$ to $150$ on Llama.

\begin{table}[h]
\centering
\caption{The knobs varied across the trained configurations.}
\label{tab:knobs}
\small
\begin{tabular}{@{}l p{0.30\linewidth} p{0.38\linewidth}@{}}
\toprule
Knob & Values tried & Meaning \\
\midrule
Witness regime & free, half held, held & share of the eligible witness pool pinned \\
$k$ schedule & ladder, $k{=}1$, reverted & decreasing, one flat $k$, or rising \\
Ladder shape & top rung $1$ to $512$, $1$ to $7$ rungs & compared as a share of the pool \\
Measurement stage & on, off & measures the leading candidates one at a time in the final steps \\
Objective & $\NIE$, $\NIE{+}\pie$, bottleneck & $\pie$ alone is out of scope \\
Prompt pairs & $100$, $150$, $200$ & $150$ on Llama only \\
\bottomrule
\end{tabular}
\end{table}

The three per-model defaults are GPT-2's \texttt{free, 200 pairs}, Qwen3's \texttt{free, interior 150, 200 pairs, measured} and Llama's \texttt{free, 142-71-43-21-7 ladder}, where the first word names the witness regime. We retrained each default under independent seeds on every task, $5$ to $35$ times per cell. Each per-cell interval in \Cref{sec:results} is a $95\%$ $t$-interval over those seeds, and each pooled interval in \Cref{fig:uncertainty} is a $t$-interval over the five per-task seed means. Cells whose first five seeds left the margin indistinguishable from zero got more seeds. That rule depends on the result it measures, so an interval that clears zero on a topped-up cell is weaker evidence than the same interval on a cell replicated to a fixed depth.

\begin{table}[h]
\centering
\caption{Settings per model. The candidate pool and the default ladder describe each model's default; the other rows summarize all of its runs.}
\label{tab:settings}
\begin{tabular}{llll}
\toprule
Setting & GPT-2 small & Qwen3-1.7B & Llama-3.1-8B \\
\midrule
Candidate pool & $144$ head writes & $448$ & $1024$ \\
Prompt pairs & $100$ or $200$ & $100$ or $200$ & $100$ or $150$ \\
Default ladder & $20$-$10$-$6$-$3$-$1$ & $150$-$67$-$33$-$8$-$1$ & $142$-$71$-$43$-$21$-$7$ \\
Steps per run & $30$ to $90$ & $30$ to $90$ & $30$ to $150$ \\
Configurations & $255$ & $159$ & $158$ \\
Seeds per default cell & $5$ to $35$ & $5$ to $10$ & $5$ to $10$ \\
\bottomrule
\end{tabular}
\end{table}

\begin{table}[h]
\centering
\small
\caption{Example factual prompts and the single change that produces each alternative prompt.}
\label{tab:tasks}
\begin{tabular}{@{}p{2.1cm}p{6.2cm}p{4.6cm}@{}}
\toprule
Task & Factual prompt $\to$ answer & Alternative prompt \\
\midrule
IOI & \textit{When Mary and John went to the store, John gave a drink to} $\to$ Mary
    & Second subject mention replaced by a third name \\
SVA & \textit{The friends} $\to$ have
    & Subject number flipped (\textit{The friend} $\to$ has) \\
Arithmetic & \textit{Q: How much is 27 plus 64? A:} $\to$ 91
    & Both operands resampled \\
Induction & A random word block, repeated: \textit{oak map \dots\ rat hat oak map \dots\ rat} $\to$ hat
    & Boundary word changed (\textit{hat} $\to$ \textit{cow}) \\
Entity tracking & \textit{The apple is in Box C. The key is in Box A. \dots\ Box C contains the} $\to$ apple
    & Same box queried, different object placed in it \\
\bottomrule
\end{tabular}
\end{table}

\section{Results in detail} \label{app:overall}

\begin{figure}[p]
\centering
\includegraphics[width=\linewidth]{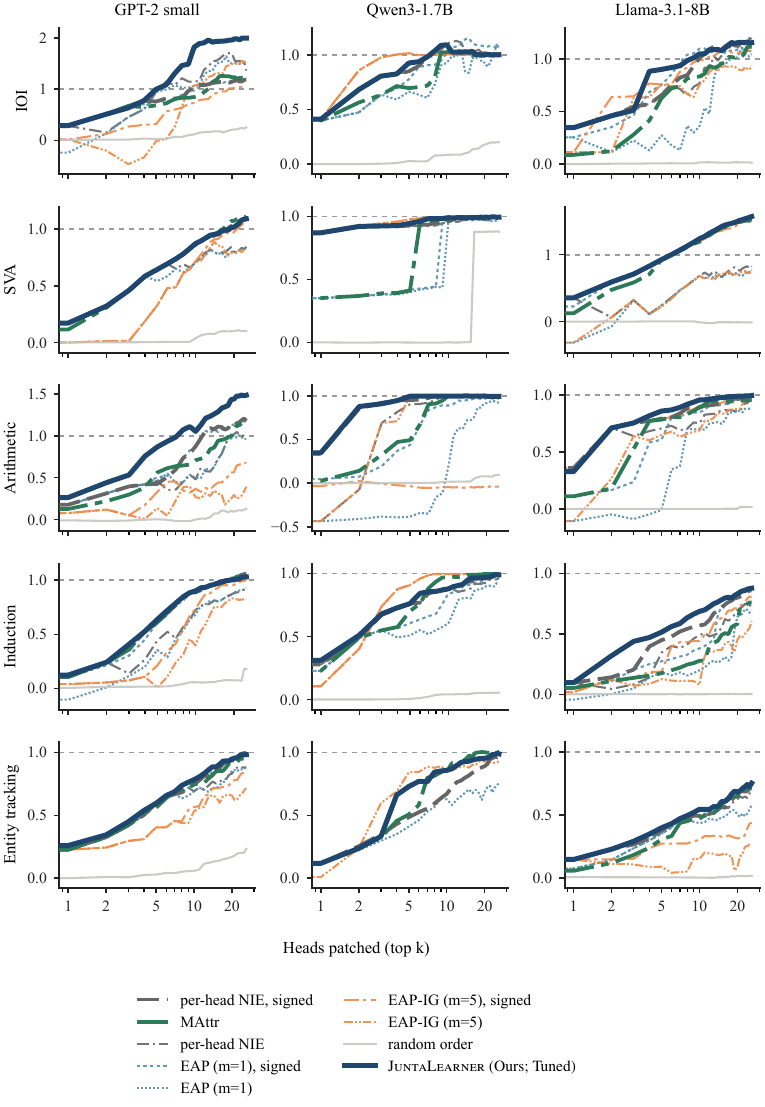}
\caption{Kill curves for each cell's best-scoring trained configuration (best configuration, for brevity) against all eight baselines. The
$x$ axis is the number of attention heads patched, log-scaled.}
\label{fig:cellcurves-kill}
\end{figure}

\begin{figure}[p]
\centering
\includegraphics[width=\linewidth]{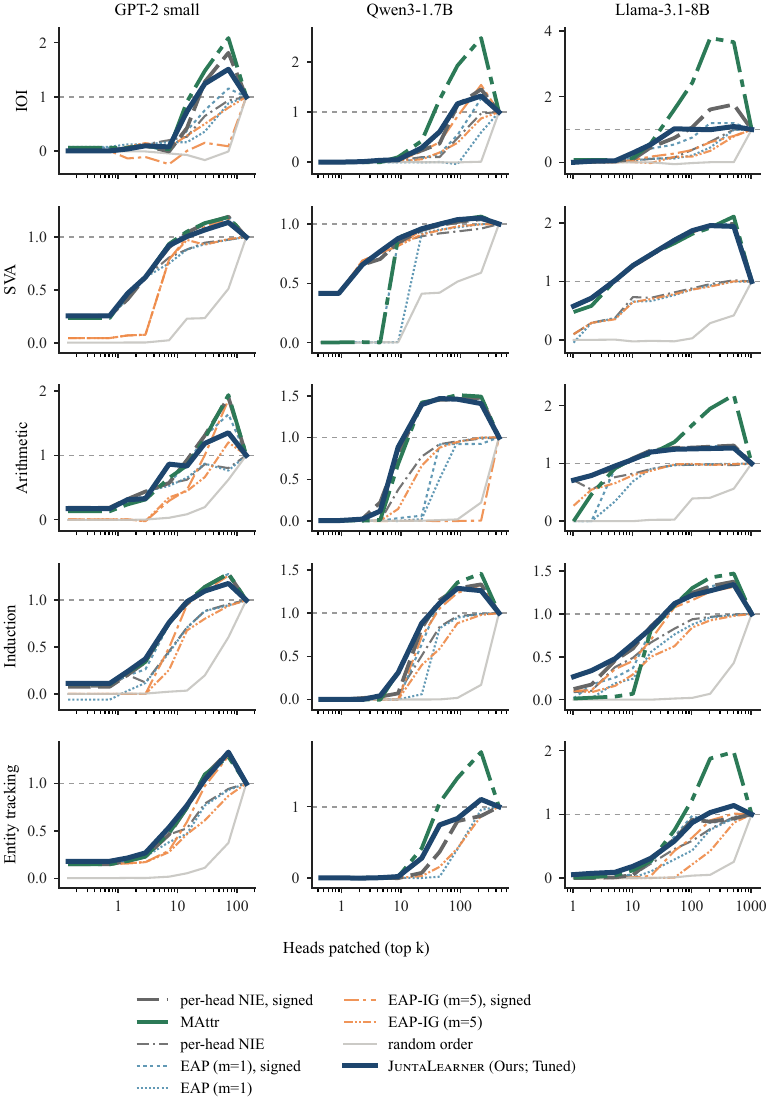}
\caption{Faithfulness curves for each cell's best configuration, selected on
$S_{\text{faith}}$ and so not always the same configuration as \Cref{fig:cellcurves-kill}.}
\label{fig:cellcurves-faith}
\end{figure}

\begin{figure}[p]
\centering
\includegraphics[width=\linewidth]{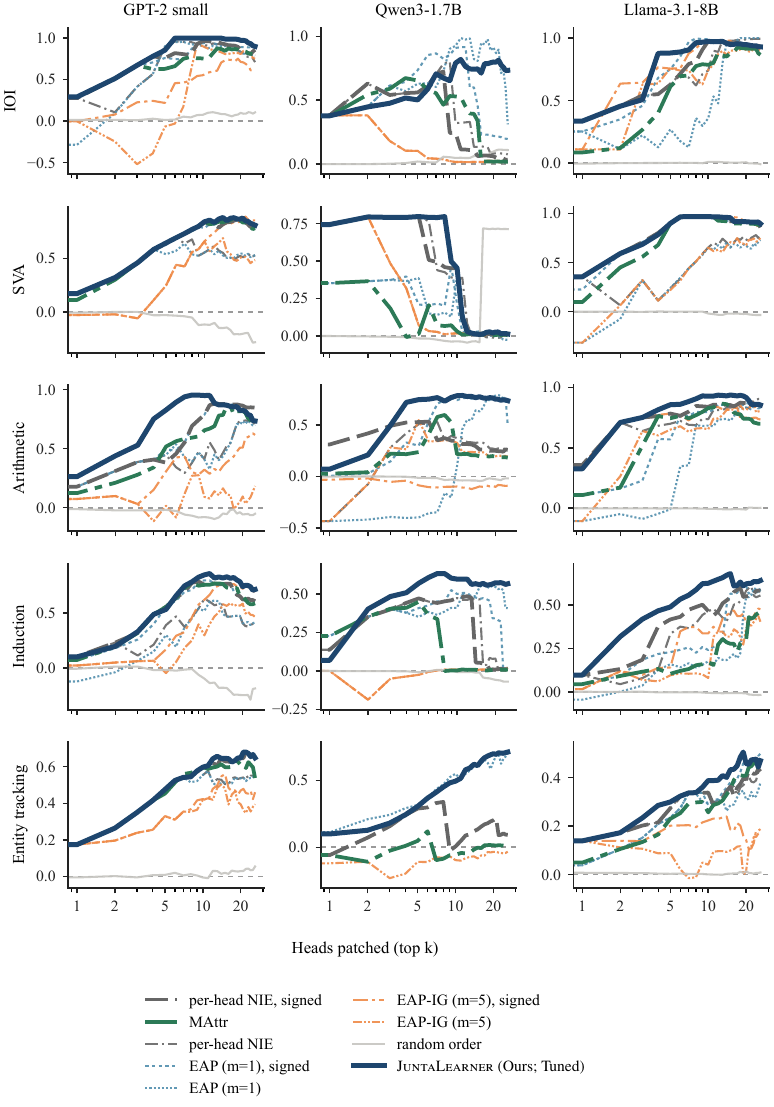}
\caption{Net curves for each cell's own best configuration. Net regularly picks a different
winner from kill and faithfulness.}
\label{fig:cellcurves-net}
\end{figure}

\Cref{fig:cellcurves-kill,fig:cellcurves-faith,fig:cellcurves-net} show,
 for each of
the fifteen cells, the best-scoring trained configuration
against all eight baselines. A reader who distrusts the aggregation can
check here whether a higher CRS corresponds to a walk they would have preferred by
eye. The winner is reselected per task and goal.

\begin{figure}[p]
\centering
\includegraphics[width=\linewidth]{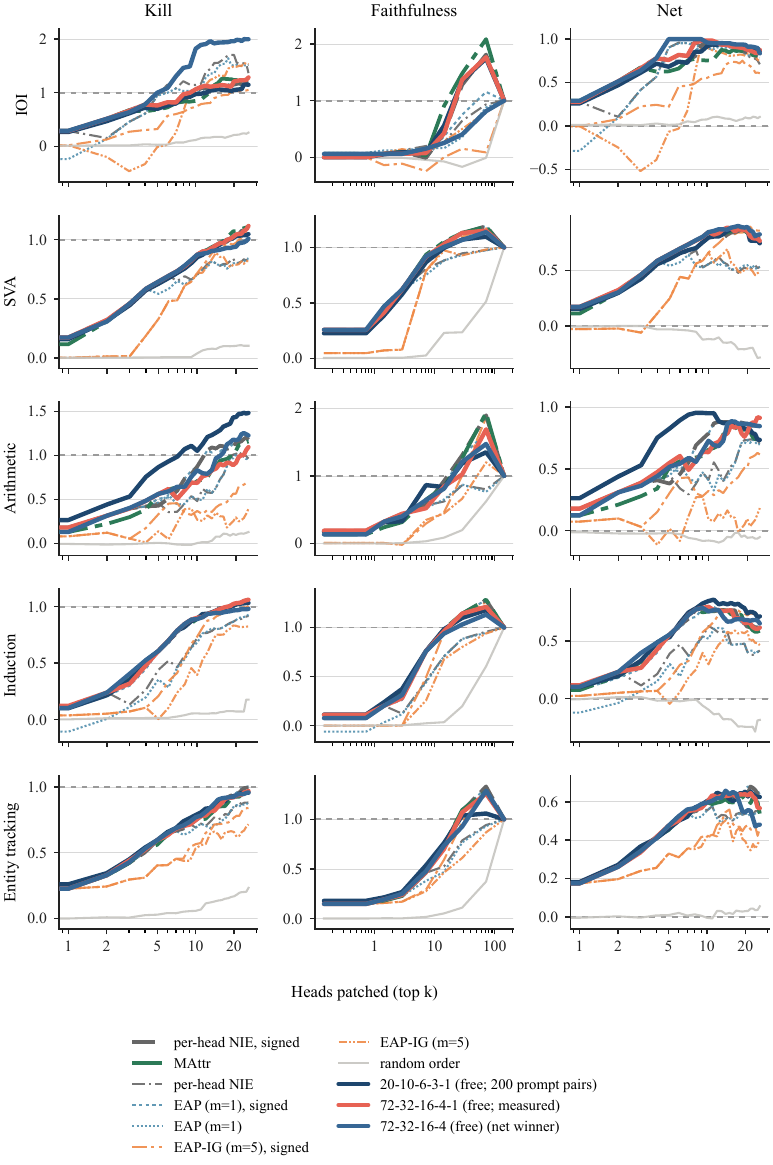}
\caption{GPT-2 small: three best settings against the baselines on all three
goals, with the CRS ranking at the right of each row.}
\label{fig:plotall-gpt2}
\end{figure}

\begin{figure}[p]
\centering
\includegraphics[width=\linewidth]{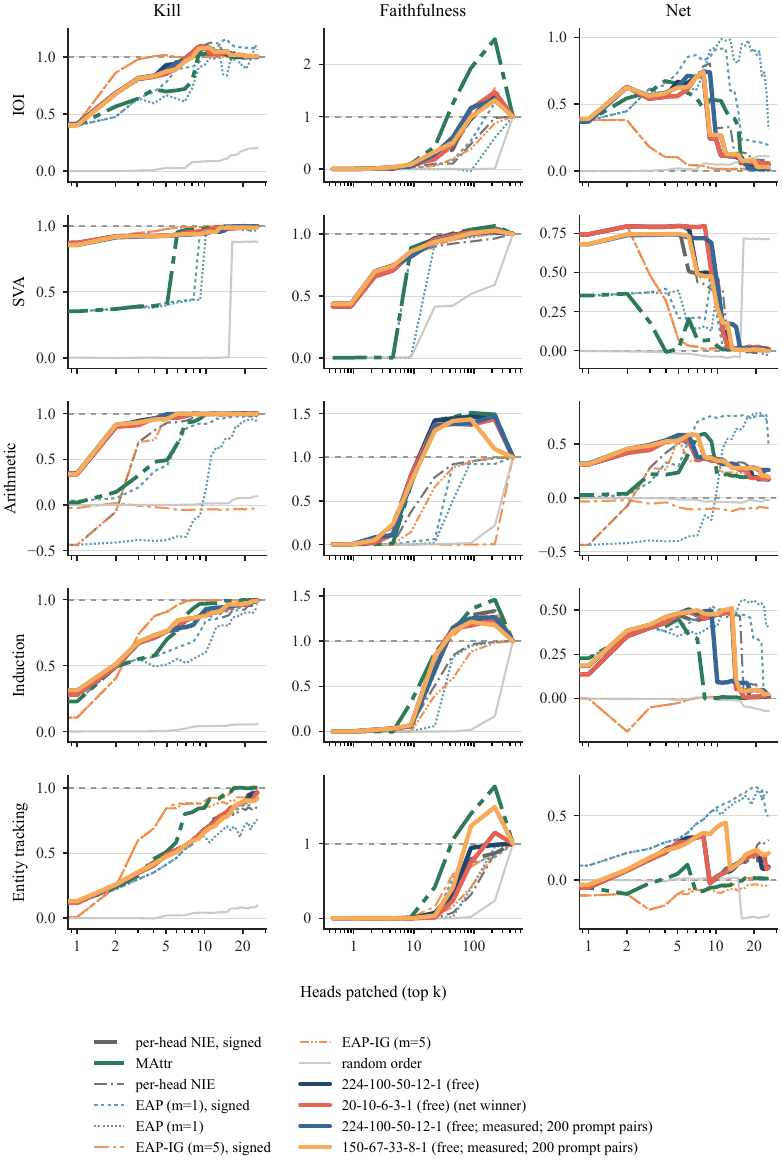}
\caption{Qwen3-1.7B, drawn as \Cref{fig:plotall-gpt2}. Its ladder shapes score within
a few thousandths of each other on kill.}
\label{fig:plotall-qwen}
\end{figure}

\begin{figure}[p]
\centering
\includegraphics[width=\linewidth]{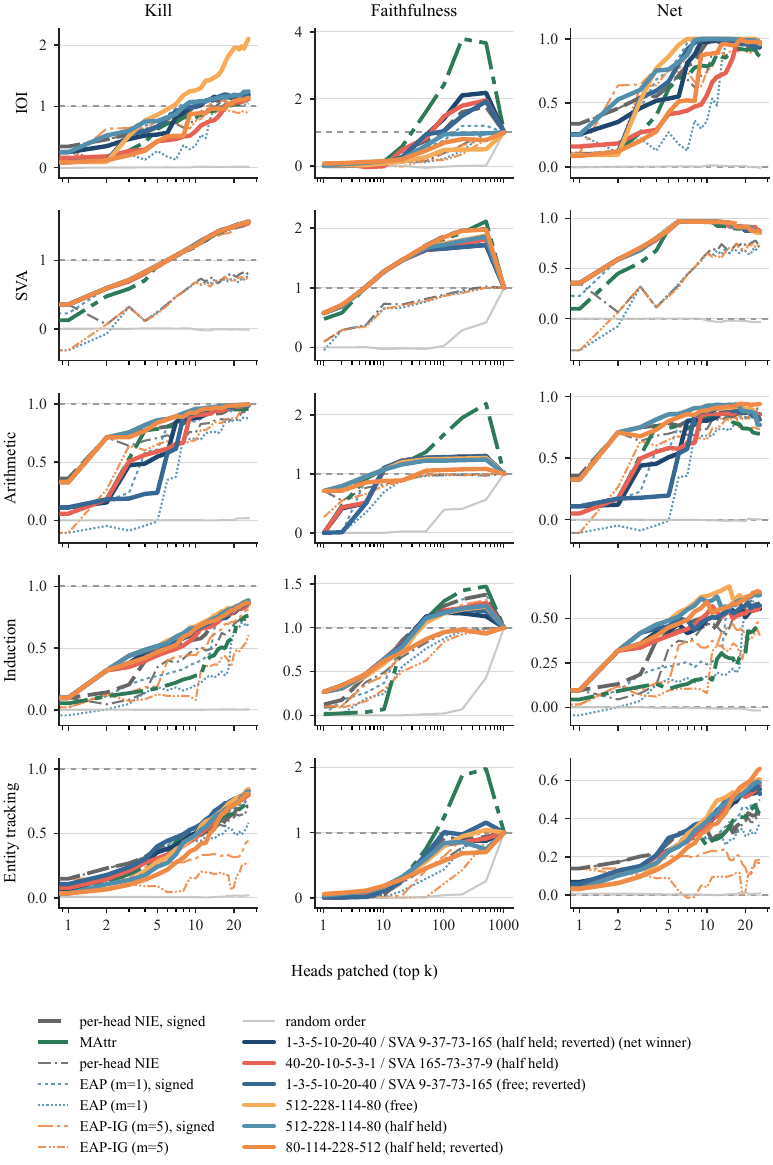}
\caption{Llama-3.1-8B, drawn as \Cref{fig:plotall-gpt2}.}
\label{fig:plotall-llama}
\end{figure}

\Cref{fig:plotall-gpt2,fig:plotall-qwen,fig:plotall-llama} put several
settings on one grid per model, with  CRS bars. The difference is that
we chose one default configuration per model, without optimizing for task-specific
performance.

\begin{figure}[p]
\centering
\includegraphics[width=0.94\linewidth]{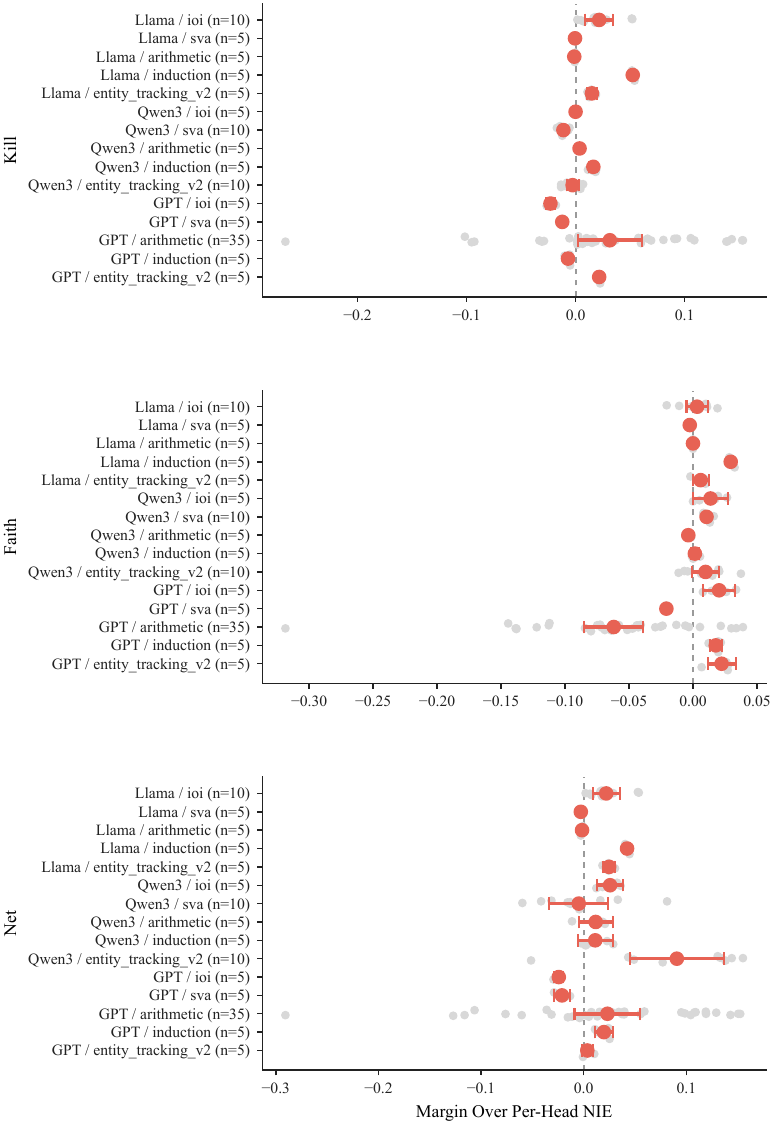}
\caption{Per-cell margins over signed per-head $\NIE$ for
model-level default configurations.}
\label{fig:percell}
\end{figure}

\begin{figure}[p]
\centering
\includegraphics[width=0.94\linewidth]{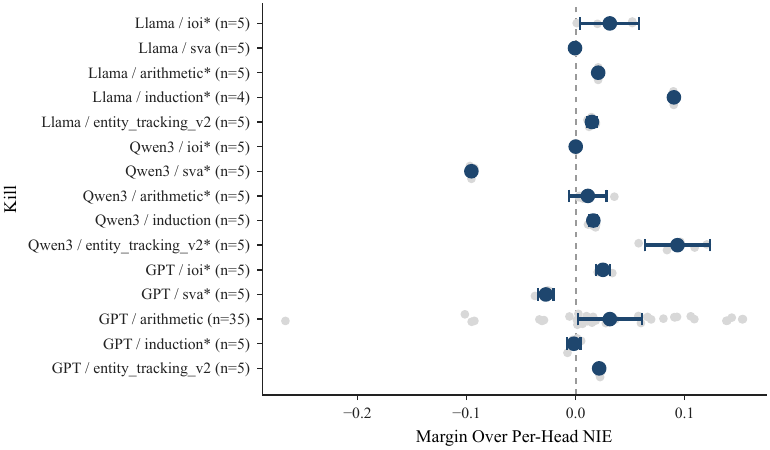}
\caption{Kill margins over signed per-head $\NIE$ for each cell's own winner, drawn as
\Cref{fig:percell}.}
\label{fig:percell-winner}
\end{figure}

Moving to uncertainty, \Cref{fig:percell} takes the per-model default,
 the single
configuration \Cref{sec:results} selects for each model,
and retrains it under several
seeds on each of that model's five tasks.
each with a $95\%$ interval.
To illustrate that defaults that work decently across
tasks can be beaten by goal and task optimized ones,
\Cref{fig:percell-winner} replaces that single defaul
t with best configuration kill-wise on that specific (model, task).

\Cref{tab:runcounts} in \Cref{app:experiments} counts the experiments actually run per
(model, task) pair, which shows that we did not extensively over-optimize per-task
configurations.

\begin{figure}[p]
\centering
\includegraphics[width=\linewidth]{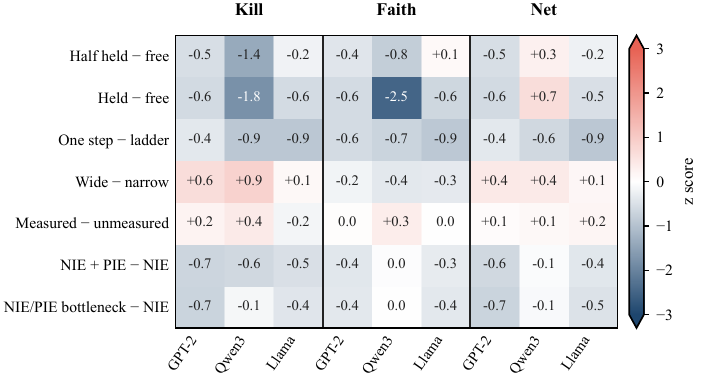}
\caption{GP-estimated effects of each knob, grouped by goal and model. Rows compare each setting while holding other settings at their observed values.
Cells give the posterior mean difference divided by its standard deviation.}
\label{fig:knobs}
\end{figure}

\begin{figure}[t]
\centering
\includegraphics[width=0.92\linewidth]{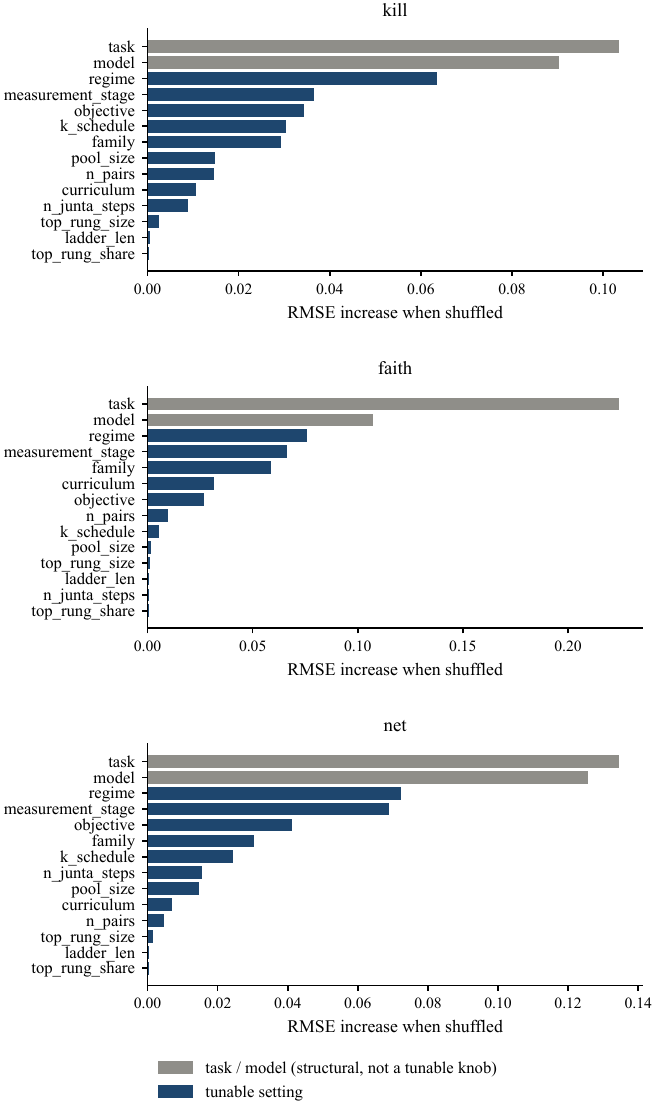}
\caption{Permutation importance over the fitted model: the increase in error when one
column is shuffled.}
\label{fig:knobimportance}
\end{figure}

\begin{figure}[p]
\centering
\includegraphics[width=\linewidth]{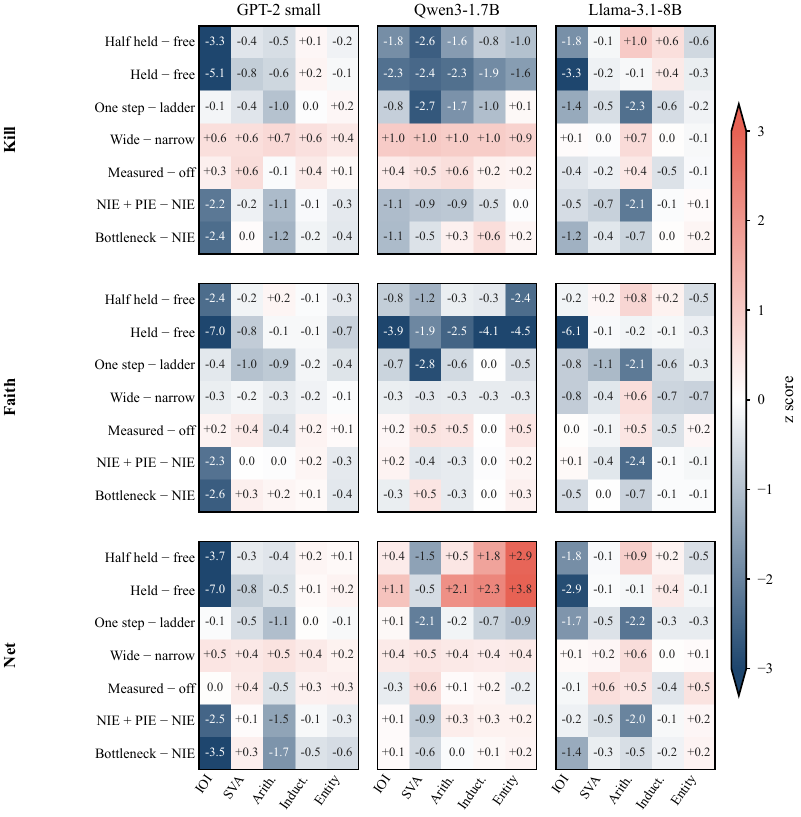}
\caption{Task-specific knob contrasts across the three goals (rows) and models
(columns). Each cell is the mean GP posterior difference divided by its standard
deviation. The signed color scale is shared across panels and clipped at $\pm 3$;
labels denote the first setting minus the second.}
\label{fig:knobs-bytask}
\end{figure}

Every contrast in \Cref{fig:knobs} and in \Cref{fig:knobs-bytask} is built using a GP
trained on the experimetnal results. For a knob $k$ with levels $a$ and $b$,
take model $m$'s own recorded training
runs $\mathcal{D}_m$. For each row $x_i \in \mathcal{D}_m$, build two copies agreeing
on every covariate except $k$, stack all $2|\mathcal{D}_m|$ of them into one input
$X$, and draw $S=300$ samples jointly from a Gaussian process fitted to that goal.
 \Cref{fig:knobs-bytask} repeats the  grid with the task fixed as well.
 \Cref{fig:knobimportance} ranks every knob by how much shuffling its column hurts the
fitted model.

\clearpage
\section{MIB Baseline and Comparison to MAttr} \label{app:mib}

In this section, we compare \textsc{JuntaLearner} with the recently released Matryoshka Attribution (MAttr, \citealp{arora2026matryoshka}), and through this comparison we assess the strengths and weaknesses of our evaluation methods alongside their incumbents, CPR and CMD on faithfulness.

\begin{figure}[htbp]
\centering
\includegraphics{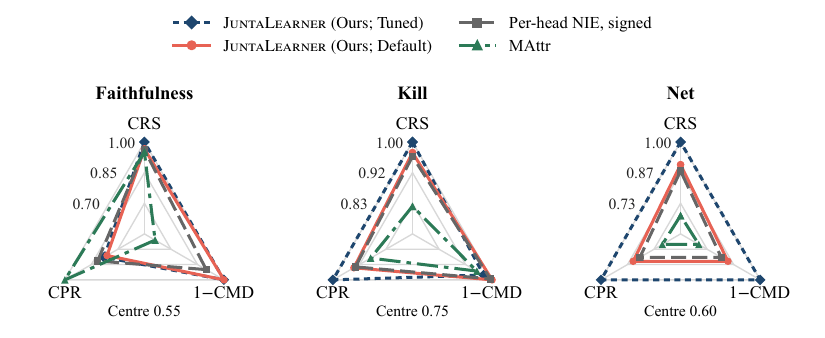}
\caption{The fifteen cells of the main experiments: \emph{ours, tuned}, \emph{ours, default},
signed per-head $\NIE$ and MAttr with its authors' published MIB recipe. Note that for readability each triangle has its own scale and that
values between goals are not directly comparable. \emph{Ours, tuned} is chosen per cell and goal by CRS, as in \Cref{tab:robustness-summary}, and scored under all three summaries. Each axis shows a method's summed score relative to the best, so outward is better on every axis; CMD appears as $1-\mathrm{CMD}$.}.
\label{fig:paper-triangle}
\end{figure}

\textbf{On our tasks, our methods lead on CRS and $1-\mathrm{CMD}$, and MAttr
leads on faithfulness CPR} (\Cref{fig:paper-triangle});  summed over all three goals, our tuned configuration and MAttr are tied on CPR (41.1 against 41.2). MAttr's profile across the three summaries is
uneven: \textbf{on faithfulness it attains the highest CPR and the lowest $1-\mathrm{CMD}$, with a CRS close
to ours}.

\begin{figure}[htbp]
\centering
\includegraphics[width=\linewidth]{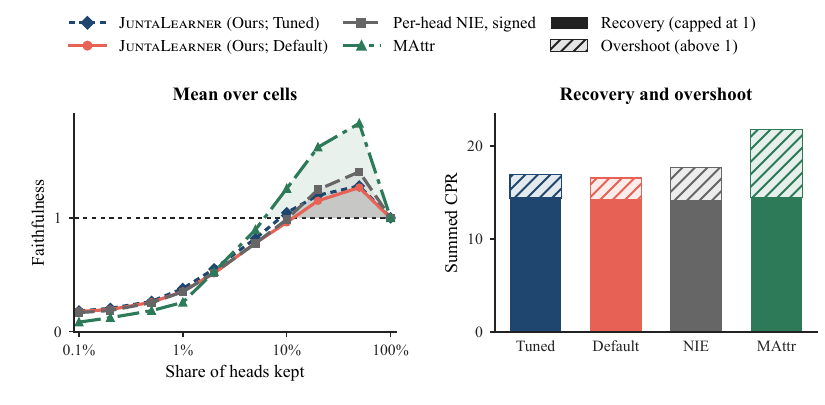}
\caption{Faithfulness for fifteen cells of the main experiments.
Left: each method's faithfulness curve, averaged over cells; shading marks where it rises
above $1$. Right: each method's faithfulness CPR, split in two components: area under the curve
capped at 1 with area above one stacked 13above.}
\label{fig:mib-overshoot}
\end{figure}

Two properties of CPR account for this. It gives substantial weight to the right-hand part
of the curve, reached after most of the relevant components have been recovered, and so
rewards late recovery as much as early detection; and it credits overshoot.
\Cref{fig:mib-overshoot} shows both effects: MAttr's mean faithfulness curve starts below
the others at small shares of the pool (left), and \textbf{most of MAttr's CPR advantage
comes from overshoot} (right).

\begin{figure}[htbp]
\centering
\includegraphics{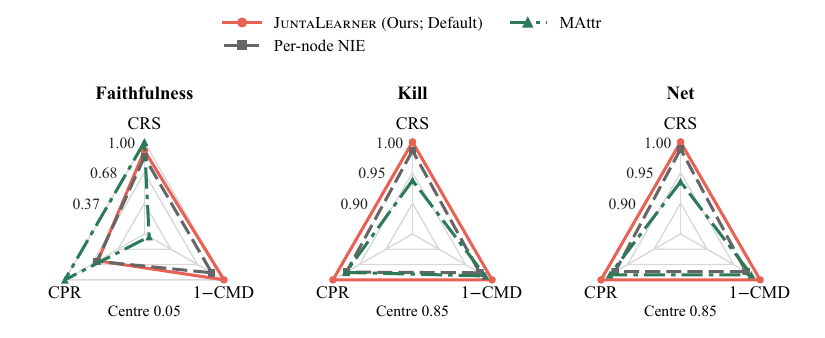}
\caption{The twelve MIB cells, public test split: \emph{ours, default} in one untuned pass, per-node $\NIE$ and MAttr's released
circuits, drawn as \Cref{fig:paper-triangle}, with faithfulness from MIB's evaluator. CMD appears as $1-\mathrm{CMD}$.}
\label{fig:mib-triangle}
\end{figure}

\begin{figure}[htbp]
\centering
\includegraphics[width=\linewidth]{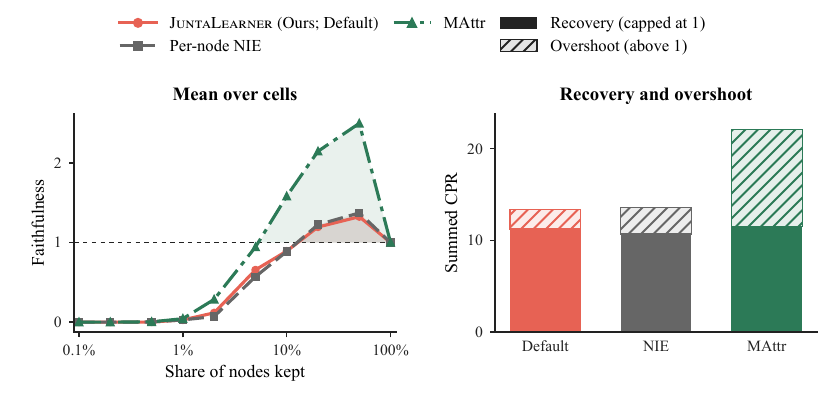}
\caption{Twelve MIB cells, drawn and computed as
\Cref{fig:mib-overshoot}.}
\label{fig:mib-overshoot-mib}
\end{figure}

MAttr seems to have been developed and evaluated mostly with the MIB's circuit track
\citep{mueller2025mib}.
\footnote{
MIB's circuit track pairs four models with six tasks in twelve cells,
scored at the node level, where a circuit is a set of attention heads, MLPs and the input embedding; its Qwen
model is Qwen2.5-0.5B. Our MIB scoring for this paper comes from the
public test split, in which we follow \citet{arora2026matryoshka}.
As for settings, our only choice on MIB data was the training objective, $\NIE$, made on
MIB validation from pilot runs of four objectives on GPT-2 IOI; a few further objective and
configuration variants on the Qwen and Llama IOI and MCQA cells served only as sensitivity
checks. Every score reported here comes from one training seed per cell.  Faithfulness comes from MIB's own evaluator. Kill and
net walk each ranking's top attention heads as in the main experiments, on up to $200$ test
pairs (as few as $17$ on MCQA and ARC), with MLPs left out, since on GPT-2 IOI patching the
first-layer MLP alone removes every task and saturates both
curves.
}
 \textbf{On MIB's public test split,
our untuned run is close to a tie with MAttr's circuits on CRS summed over the three goals,
ahead on $1-\mathrm{CMD}$ and behind on CPR}. As we also care about the other evaluation axes,
we include also triangles that correspond to them (\Cref{fig:mib-triangle}). \textbf{On kill and net, summed
over cells, our untuned defaults beat MAttr on all three summaries.}  Moreover, the CPR dominance for MAttr comes to a large extent
from overshoot here too. \textbf{Most of MAttr's CPR lead
on MIB comes from overshoot.} (\Cref{fig:mib-overshoot-mib}). That being said, notice how the mean MAttr
for faithfulness here climbs a bit faster than ours, and so \textbf{CRS rightly
 prefers MAttr's released circuits over our untuned default for faithfulness on MIB}.

\begin{figure}[p]
\centering
\includegraphics[width=\linewidth]{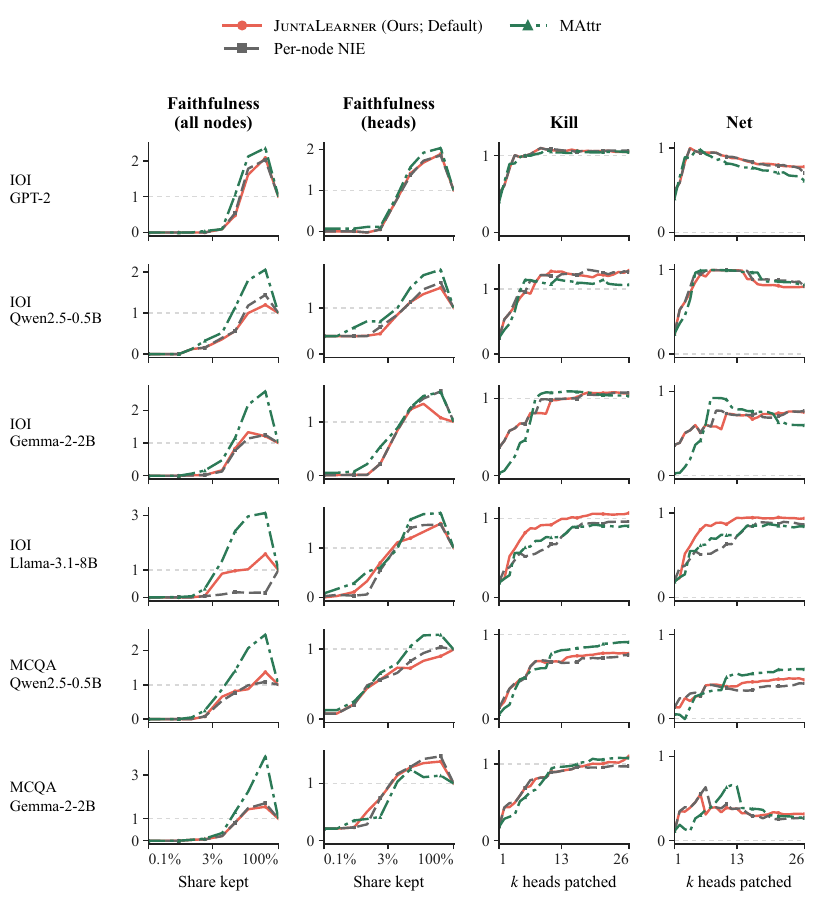}
\caption{MIB public test, the four IOI cells and two of the three MCQA cells. Columns: MIB faithfulness over all nodes and
faithfulness over heads (share of the pool kept), then kill and net (top $k$ heads
patched). Dashed grey marks $f=1$ and kill $d=1$ ($d$ is the share of the full-pool effect
removed), the ideal values, and net $=0$, where a
circuit's damage to the target equals its mean damage to the controls; the ideal net is
$1$.}
\label{fig:mib-curves-a}
\end{figure}

\begin{figure}[p]
\centering
\includegraphics[width=\linewidth]{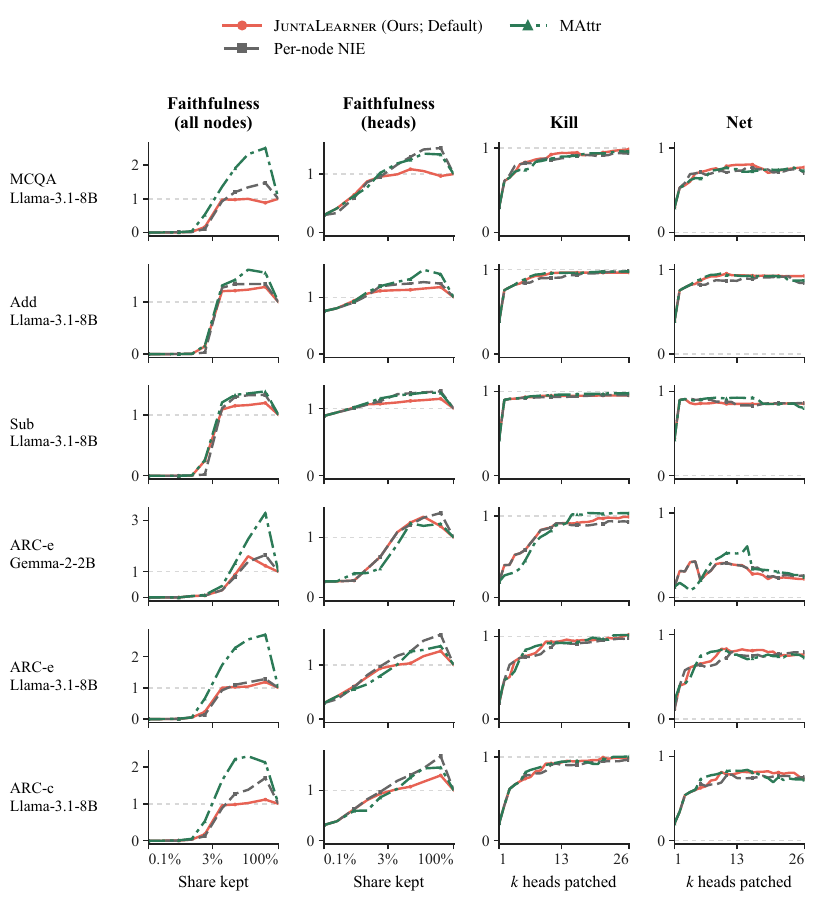}
\caption{MIB public test, the remaining cells, drawn as \Cref{fig:mib-curves-a}.}
\label{fig:mib-curves-b}
\end{figure}

MAttr is a strong method, and the two methods are trained in opposite directions, which
may explain some of the uneven picture. MAttr's released circuits are optimized for what MIB
faithfulness explicitly scores: to keep a set of components and patch everything else.
Ours are trained by patching the set itself, which is closer
to what is scored by kill.  Sums over cells can hide how
these differences play out on individual cells, so \Cref{fig:mib-curves-a,fig:mib-curves-b}
provide every curve behind them, one row per cell (MIB's faithfulness, faithfulness over
heads, kill and net).

\end{document}